\documentclass[10pt,twocolumn,letterpaper]{article}

\usepackage[pagenumbers]{cvpr} 

\usepackage[toc,page]{appendix}
\usepackage{comment}
\usepackage[dvipsnames]{xcolor}
\usepackage{graphicx}
\usepackage[ruled]{algorithm2e}  
\usepackage{booktabs}  
\usepackage[pagebackref,breaklinks,colorlinks,citecolor=citecolor]{hyperref}

\usepackage{listings}
\usepackage{balance}
\usepackage{tabularx}
\usepackage[none]{hyphenat}
 
\def\paperID{48} 
\def\confName{3DV\xspace}
\def\confYear{2024\xspace}

 \newcommand{\papertitle}{Text to Animatable Digital Avatars}

\newcommand{\modelname}{\textcolor{black}{TADA}\xspace}
\newcommand{\projectURL}{\href{https://tada.is.tue.mpg.de/}{\tt\textit{tada.is.tue.mpg.de}}}
\definecolor{citecolor}{HTML}{0071bc}

\newcommand{\Eref}[1]{Eq.~(\ref{#1})}
\newcommand{\Fref}[1]{Fig.~\ref{#1}}
\newcommand{\Sref}[1]{Sec.~\ref{#1}}

\newcommand{\tref}[1]{Tab.~\ref{#1}}

\newcommand{\fref}[1]{Fig.~\ref{#1}}
\newcommand{\sref}[1]{Sec.~\ref{#1}}

\definecolor{jtcolor}{RGB}{0,0,255}

\definecolor{todocolor}{RGB}{255,0,00}

\newcommand{\qheading}[1]{\noindent\textbf{#1}.}
\newcommand{\mheading}[1]{\medskip\noindent\textbf{#1}.}

\newlength\savewidth\newcommand\shline{\noalign{\global\savewidth\arrayrulewidth
  \global\arrayrulewidth 1pt}\hline\noalign{\global\arrayrulewidth\savewidth}}
\author{
    Tingting Liao\textsuperscript{1}$^{*}$,
    Hongwei Yi\textsuperscript{2}$^{*}$,
    Yuliang Xiu\textsuperscript{2},
    Jiaxiang Tang\textsuperscript{3},
    Yangyi Huang\textsuperscript{4}\\
    Justus Thies\textsuperscript{2},
    Michael J. Black\textsuperscript{2}
    \\
    {\normalsize\textsuperscript{1}Mohamed bin Zayed University of Artificial Intelligence
    \quad
    \textsuperscript{2}Max Planck Institute for Intelligent Systems}
    \\
    {\normalsize\textsuperscript{3}Peking University
    \quad
    \textsuperscript{4}State Key Lab of CAD \& CG, Zhejiang University}
    \\   
    {\tt\small tingting.liao@mbzuai.ac.ae, tjx@pku.edu.cn, huangyangyi@zju.edu.cn}\\
    {\tt\small \{hongwei.yi, yuliang.xiu, justus.thies, black\}@tuebingen.mpg.de}\\
}

\title{\modelname! \papertitle} 

\begin{document}

\newcommand{\teaserCaption}{
With only text descriptions as input, \modelname generates high-fidelity 3D avatars with lifelike texture and detailed geometry, including high-resolution faces. 
Accurate alignment of texture and geometry, together with an underlying SMPL-X representation, enables expressive animation.
\modelname also supports applications such as virtual try-on and personalized editing using text.
}

\twocolumn[{
    \renewcommand\twocolumn[1][]{#1}
    \maketitle
    \vspace{-0.7 cm}
    \centering
    \begin{minipage}{1.00\textwidth}
        \centering 
        \includegraphics[trim=000mm 000mm 000mm 000mm, clip=False, width=\linewidth]{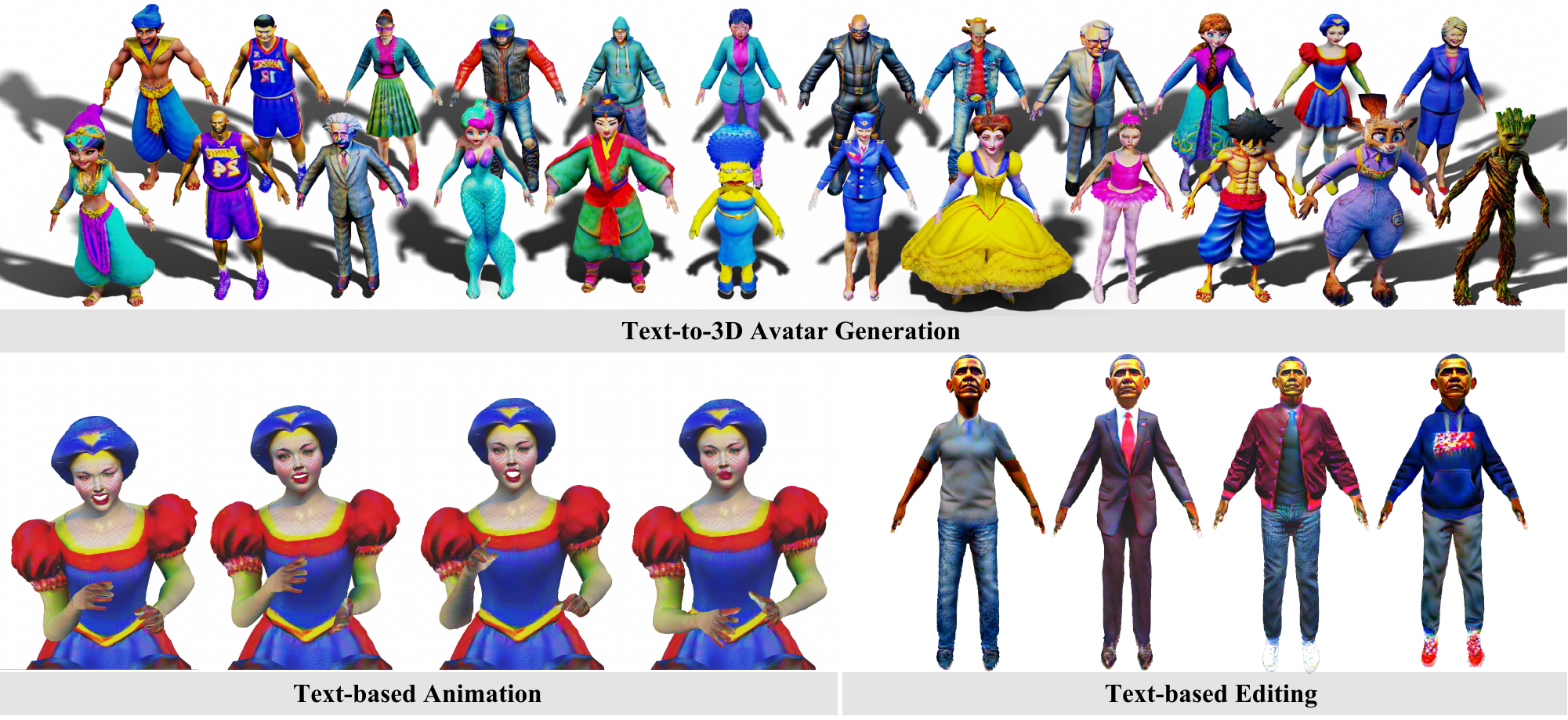}        
    \end{minipage}
    \vspace{-0.5 em}
    \captionsetup{type=figure}
    \captionof{figure}{\teaserCaption}
    \label{fig:teaser}
    \vspace{1.0em}
}]

\def\thefootnote{*}\footnotetext{denotes equal contribution.}

\begin{abstract}
We introduce \modelname, a simple-yet-effective approach that takes textual descriptions and produces expressive 3D avatars with high-quality geometry and lifelike textures, that can be animated and rendered with traditional graphics pipelines.
%
Existing text-based character 
generation methods are limited in terms of geometry and texture quality, 
and cannot be realistically animated due to the misalignment between the geometry and the texture, particularly at face region.
To address these limitations, \modelname leverages the synergy of a 2D diffusion model and a parametric body model.
Specifically, we derive a high-resolution upsampled SMPL-X with displacement layer and a texture map, and use hierarchical rendering with score distillation sampling (SDS) to create high-quality, detailed, holistic 3D avatars from text.
To ensure alignment between the geometry and texture, we render normals and RGB images of the generated character and exploit their latent embeddings during SDS optimization process.
We further drive the face of character with multiple expressions during optimization, ensuring that its semantics remain consistent with the original SMPL-X model, for realistic animation with semantic alignment. 
Both qualitative and quantitative evaluations show that \modelname significantly surpasses existing approaches.
\modelname enables large-scale creation of digital characters ready for animation and rendering, while also allows for text-guided editing. 
The code will be public for research purposes at \projectURL.
\end{abstract}
    
\section{Introduction}
\label{sec:introduction}
Digital avatars are a foundation for applications in augmented and virtual reality, immersive telepresence~\cite{li2020monoport,li2020monoportRTL, zheng2023avatarrex, saito2020pifuhd, xiu2023econ}, virtual try-on~\cite{zhu2022registering, santesteban2021ulnefs, santesteban2021garmentcollisions}, and video games~\cite{zhu_2020_eccv_nba, deepcap, zheng2021deepmulticap}.
Creating high-quality and expressive 3D avatars is challenging since the geometry and appearance of the character has to be modelled under a variety of different poses.
Traditional pipelines used in the entertainment industry often use sophisticated multi-view capture studios~\cite{ma2008facial, bickel2007multi, joo2015panoptic} to create complex 3D models. Manual processes like cleaning and rigging the scans make creating an animatable character time-consuming and expensive.
While there is recent progress on automatic learning-based body reconstruction from single image \cite{saito2019pifu, saito2020pifuhd, xiu2023econ, xiu2022icon, huang2022elicit, li2020monoport, liao2023car,yang2023dif}, or sparse images \cite{shao2022diffusionstereo}, such methods are limited to real humans, fail on fictional characters, and are hard to edit and control.
Thanks to the rapid progress on Large Language Models \cite{raffel2020exploring, brown2020language} and Diffusion Models \cite{welling2011bayesian, sohl2015deep, ho2020denoising, song2019generative, song2020denoising}, recent work has shown that text-to-image models \cite{ramesh2021zero, rombach2022high} can be combined with differentiable neural 3D scene representations such as DeepSDF \cite{park2019deepsdf}, NeRF \cite{mildenhall2020nerf} and \textsc{DMTet} \cite{shen2021deep} to generate realistic 3D models solely from textual descriptions. 
However, these methods have many limitations. 
The generated objects or characters are often rigid and lack of animation \cite{lin2022magic3d, text2mesh, richardson2023texture, TextDeformer, chen2023fantasia3d}, they have difficulty in producing high-quality realistic 3D avatars in terms of geometry and texture~\cite{hong2022avatarclip}, or the characters are incompatible with traditional CG workflows (NeRF based~\cite{cao2023dreamavatar, kolotouros2023dreamhuman, poole2022dreamfusion}).

\begin{figure}[t]
\centering
\includegraphics[width=\linewidth]{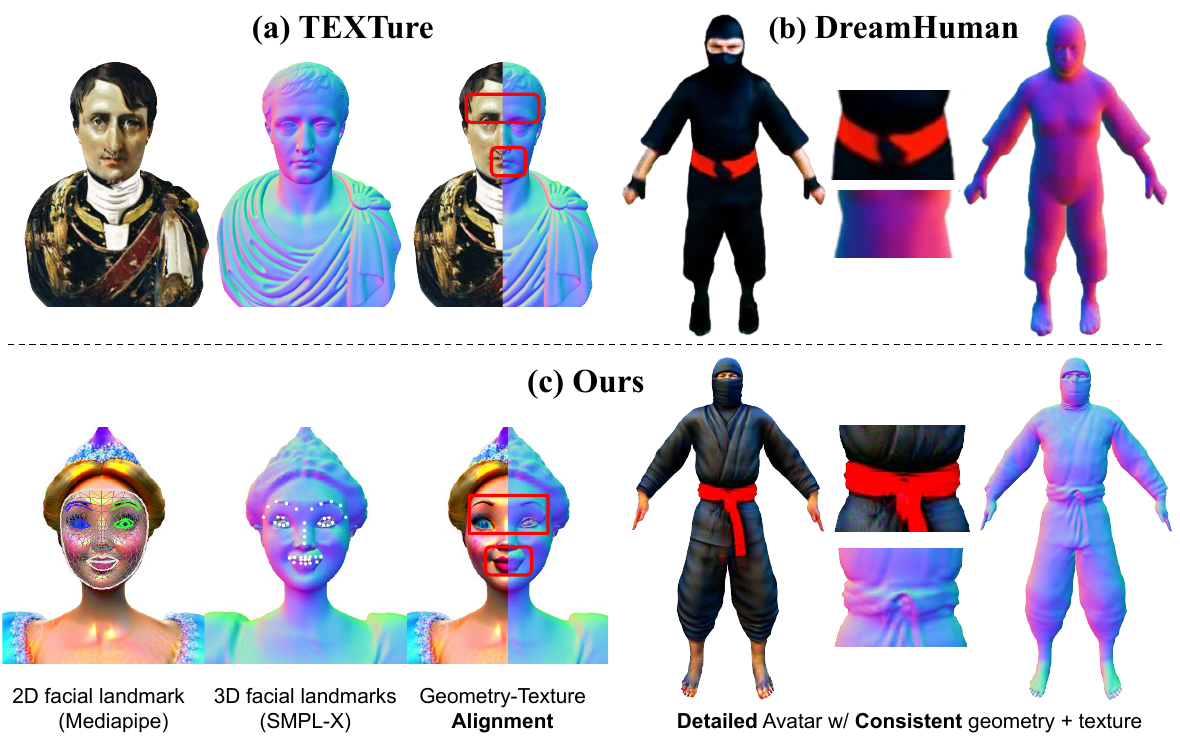} 

\vspace{-0.5 em}
\caption{
    Compared with other existing methods \cite{kolotouros2023dreamhuman, poole2022dreamfusion}, our method can generate high-quality 3D avatars with well-aligned geometry and texture that is consistent with SMPL-X,  enabling animation and rendering with existing graphics pipelines. 
}
\label{fig:illustration}
\end{figure}

Here we address these limitations with \modelname, illustrated in Fig.~\ref{fig:teaser}.
Since our goal is animatable avatars that are compatible with existing rendering engines, we build upon the SMPL-X body model \cite{SMPL-X:2019}.
SMPL-X, however, only represents a realistic, minimally clothed human body shape.
Our goal is to create diverse avatars with a wider variety of body shapes and textures.
Like recent work that generates avatars from text, we use Score Distillation Sampling (SDS)~\cite{poole2022dreamfusion} but do so in several novel ways.
Specifically, we make three key contributions.
%
%
%
%
%
%
%
(i) First, we devise a hierarchical optimization of a hybrid mesh representation which is based on a  subdivided version of SMPL-X~\cite{SMPL-X:2019} with additional learnable displacements and a texture map.
To produce high-quality details, especially on the face region, we perform hierarchical optimization over hierarchically rendered images with different focal lengths, where the entire body, or only specific parts, are visible.
(ii) Second, existing text-to-3D object methods~\cite{poole2022dreamfusion, chen2023fantasia3d} suffer from inconsistent alignment between the reconstructed geometry and texture (see \fref{fig:illustration}), as they evaluate the texture and geometry individually.
This makes animation of the resulting avatars infeasible.
%
%
%
%
(iii) Third, we want the generated character to be semantically consistent with SMPL-X so that it can be easily animated; that is, body parts and vertices on output avatars correspond to the same body parts and vertices on SMPL-X.
To that end,
we introduce animations throughout the optimization process. Specifically, we deform the generated character in each optimization step by sampling predefined SMPL-X body poses and facial expressions. This ensures that our generated characters can be animated accurately and coherently, as depicted in \fref{fig:illustration} (c).
Notably, once optimization is finished, our generated characters can be animated with any novel set of SMPL-X parameters.
Especially, combined with existing text to motion generation~\cite{shafir2023human, tevet2023human} or text-to-audio-to-motion methods~\cite{tts, yi2022talkshow}, we can animate the generated characters to interact with the scenes or communicate with others.
This paves the way towards creating virtual 3D worlds with animatable digital avatars fully from text.


\medskip
\noindent
In summary, with \modelname we propose a user-friendly tool for avatar creation and editing, that can be solely controlled by the textual input and is fully compatible with traditional graphics pipelines.
The output model is graphics-ready because the underlying model is SMPL-X with displacements and a texture map.
Our method can generate realistic iconic celebrities, customized humans, and cartoon characters.
We validate our contributions with ablation studies, show qualitative comparisons to the state of the art, and conduct a user study that quantifies the performance of our method on the task of high-quality 3D avatar generation.

\begin{figure*}[ht!]
\centering
\vspace{-1.0 em}
\includegraphics[width=\linewidth]{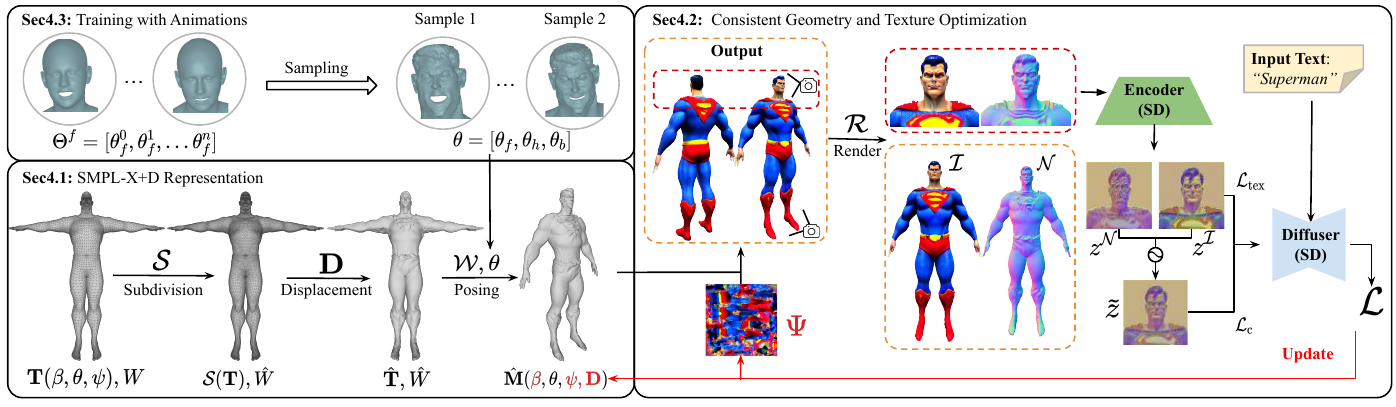}
\vspace{-1.5 em}
\caption{
\textbf{Overview}.
Initialized by a SMPL-X body $\mathbf{T}(\beta, \theta, \psi)$ with skinning weights $W$, we subdivide the body to obtain a denser mesh $\mathbf{S}(\mathbf{T})$ and add personalized displacements $\mathbf{D}$ to it. 
The personalized mesh $\hat{\mathbf{T}}$ is transformed into the posed space denoted as $\hat{\mathbf{M}}$ using randomly sampled expressions and poses from an animation database. 
In each optimization step, the expressions and poses are changed and $\hat{\mathbf{M}}$ is rendered under a novel view.
Based on the rendered RGB $\mathbf{I}$ and normal image $\mathbf{N}$, the geometry and texture of the mesh are simultaneously optimized by a Score Distillation Sampling (SDS) loss.
}  
\label{fig:pipeline}
\vspace{-1.0 em} 
\end{figure*}

\section{Related Work}
\label{sec:related_work}

%


%
Recently, there has been rapid progress on extending text-to-2D-image generation methods~\cite{Karras2019stylegan2,fu2022styleganhuman, ruiz2022dreambooth} to text-to-3D-content generation~\cite{metzer2022latentnerf, poole2022dreamfusion, wang2023prolificdreamer}.
Here, we discuss the most relevant text-to-3D-content generation methods, while focusing primarily on text-to-3D-avatar generation (both human and anime characters).
\qheading{Text-to-3D-Content Generation} 
%
%
The successes of Text-to-Image (T2I) generative modeling \cite{rombach2022high, saharia2022photorealistic,dalle2} have sparked a surge of interest in the field of text-to-3D generation \cite{sanghi2022clipforge, wang2022clipnerf, chen2022tango, xu2022dream3d}. 
Despite progress, effectively describing and controlling 3D properties of an object using language, while ensuring coherence in the three-dimensional space, remains a challenge. 
One line of work \cite{text2mesh, mohammad2022clipmesh, jain2021dreamfields} utilizes CLIP-space similarities to guide shape and texture optimization.
However, these methods often fail to generate convincing and realistic 2D renderings.
CLIP-based optimization can be combined with a generative appearance model to improve the  quality, as shown in CLIPFace~\cite{aneja2023clipface}.
However, this requires learning a GAN-model for the 3D appearance, which is challenging for full-body avatars that can vary from real humans to cartoon characters.
To circumvent the training of a 3D generative model and the problem of missing datasets, recent publications~\cite{richardson2023texture, poole2022dreamfusion, lin2022magic3d, chen2023fantasia3d} make significant strides by leveraging the power of score distillation sampling (SDS) \cite{rombach2022high} derived from 2D text-to-image diffusion models to create content from textual descriptions by optimizing a 3D representation.
TEXTure~\cite{richardson2023texture} takes a mesh as input and only optimizes the texture map based on a given text prompt. 
In contrast, DreamFusion~\cite{poole2022dreamfusion} optimizes a Neural Radiance Field (NeRF) \cite{mildenhall2020nerf}  to represent the 3D content in terms of a density and radiance field.
It can generate 3D models of arbitrary (fictional) objects.
However, it faces challenges due to slow optimization of NeRF and low-resolution image space supervision, resulting in long processing times and low-quality 3D models. 
To overcome these limitations, Magic3D~\cite{lin2022magic3d} introduces a two-stage optimization framework, using NeRF in the first stage and a textured mesh in the second stage.  
Fantasia3D~\cite{chen2023fantasia3d} extends this to generate 3D meshes by disentangling geometry and texture, and optimizes them separately.  
All these methods focus on general, static, object/scene generation; they output is not animation-ready, which is necessary for 3D character creation. 

\qheading{Text-to-3D-Avatar Generation} 
Several methods generate 3D head avatars from text~\cite{zhang2023dreamface, han2023headsculpt, Gao_2023_SIGGRAPH, zhao2023zero, wang2023rodin}.
In contrast, we focus on generating full-body characters including the detailed face.
AvatarCLIP~\cite{hong2022avatarclip} leverages NeuS~\cite{wangNeuSLearningNeural2021} and the SMPL-X model with a CLIP-guide loss to facilitate the generation of avatars.
%
Similarly, DreamAvatar~\cite{cao2023dreamavatar} utilizes the shape parameters from SMPL as a prior to learn a NeRF-based color field.   
DreamHuman~\cite{kolotouros2023dreamhuman} leverages imGHUM~\cite{alldieck2021imghum} as a prior, which represents a signed distance field conditioned on pose and shape parameters, to learn a NeRF of the human.
However, the NeRF representation remains problematic due to its relatively low geometry and appearance quality, and it is not compatible with traditional graphics workflows, especially for animation. 
%
In the domain of explicit representations, Text2Mesh~\cite{text2mesh} and Chupa~\cite{kim2023chupa} employ vertex displacement on a predefined mesh template. Nonetheless, the inherent limitation of fixed topology poses challenges in accurately generating diverse character shapes.
In contrast, our approach jointly optimizes the shape, expression, and displacement. Thus, the generated characters exhibit superior quality, can be easily animated with SMPL-X motions, and seamlessly integrate into existing CG rendering and animation workflows.
%
 



\section{Preliminaries}
\label{sec:preliminaries}

\noindent \textbf{SMPL-X}~\cite{SMPL-X:2019}
is an animatable parametric 3D body model that consists of the human body, hands and face.
It has $N=10,475$ vertices and $K=54$ joints.
Given the shape $\beta$, pose $\theta$ (including body joints pose $\theta_{b}$, jaw pose $\theta_{f}$ and finger pose $\theta_{h}$) and expression $\psi$ parameters, SMPL-X models the human body as $\mathbf{M}(\beta, \theta, \psi)$:
\begin{align}
\begin{split}
    \mathbf{M}(\beta, \theta, \psi) &= \mathcal{W}(\mathbf{T}(\beta, \theta, \psi), J(\beta), \theta, W) \\
    \mathbf{T}(\beta, \theta, \psi) &= T + B_s(\beta) + B_e(\psi) + B_p(\theta) ,
\label{equ:smplx}
\end{split}
\end{align}
where $T$ is a mean shape template, $B_s, B_e$ and $ B_p$ are shape, expression and pose blend shapes, respectively.
$\mathcal{W}$ is the linear blend-skinning function transforming $\mathbf{T}(\beta, \theta, \psi)$ to the target pose $\theta$, with the skeleton joints $J(\beta)$ and skinning weights $W \in \mathbb{R}^{N\times K}$.
%

\noindent \textbf{Score Distillation Sampling}~\cite{poole2022dreamfusion}
has been proposed in DreamFusion to utilize a pre-trained 2D diffusion model to optimize the parameters $\eta$ of a 3D model, given a text $y$ as input.
%
Given the diffusion model $\phi$ with the noise prediction network $\hat{\epsilon}_{\phi}(x_t; y, t)$, SDS optimizes parameters $\eta$ by directly minimizing the injected noise $\epsilon$ added to the rendered images $x = g(\eta)$ and the predicted noise:
\begin{equation}
    \nabla_{\eta} \mathcal{L}_{SDS}(\phi, x) = E_{t,\epsilon} \left[w(t)(\hat{\epsilon}_{\phi}(x_t; y, t) - \epsilon) \frac{\partial x}{\partial \theta} \right] ,
\end{equation}
where $g(\eta)$ denotes the differentiable rendering of the 3D model parameterized by $\eta$, $x_t$ is the noised image, and $w(t)$ is a weighting function that depends on the noise level $t$.

\section{Method}
\label{sec:method} 
Given an input text prompt, \modelname aims to generate a high-fidelity animatable full-body avatar.  
As illustrated in \Fref{fig:pipeline}, our method initializes the 3D avatar with upsampled SMPL-X, which is parameterized with shape, pose, and expression parameters.
Based on it, learnable displacements are incorporated, resulting in a ``clothed'' avatar with increased density (\Sref{sec:smplx_representation}).
Then, we optimize the 3D character with consistent geometry and texture using SDS losses that considers both the rendered normal and RGB images in the latent space (\sref{sec:geo_tex_learning}). 
To encourage semantic consistency with the SMPL-X, we sample different gestures and expressions during training (\sref{sec:multi_expression_training}).
This enables the future animation using the SMPL-X pose and expression space.

\begin{figure*}[!ht]
\centering
\includegraphics[width=\linewidth]{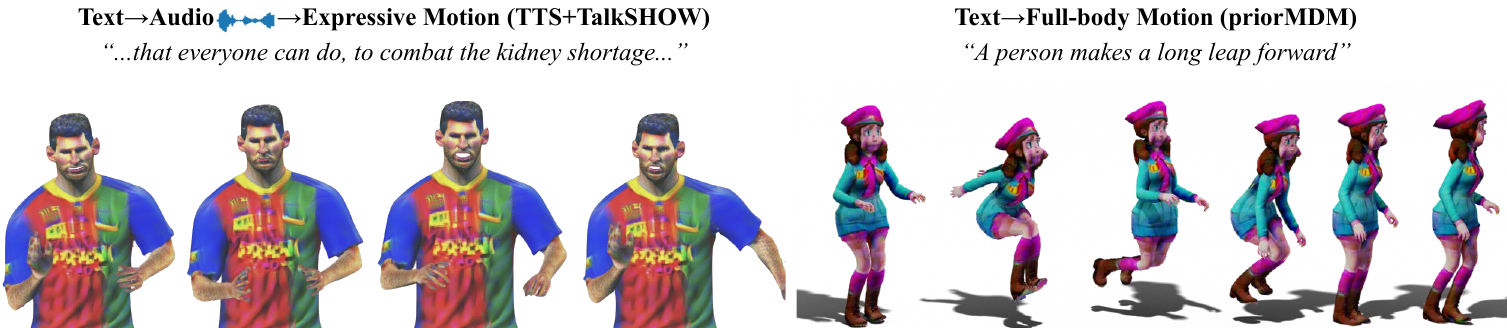}  
\caption{
\modelname enables holistic animation over the face, body and hands.
We show animation examples of the avatars ``\textit{Lionel Messi}" and ``\textit{Mabel Pines in Gravity Falls}" using expressions and body poses from TalkSHOW \cite{yi2022talkshow} (with TTS~\cite{tts}) and priorMDM \cite{shafir2023human}, respectively. 
}
\vspace{-1.0 em}
\label{fig:animation}
\end{figure*}

\subsection{SMPL-X+D Representation}  
\label{sec:smplx_representation}
%
%
%
%
\modelname adopts an SMPL-X+D to model animatable clothed avatars. 
The learnable displacement (D) accounts for personalized details that are independent of pose, shape, and expression. 
%
To generate a high-quality character with a detailed face, we apply a partial mesh subdivision on the original SMPL-X model, which is adapted as (\Eref{equ:smplx}): 
\begin{align}
\begin{split}
    \hat{\mathbf{M}}(\beta, \theta, \psi, \mathbf{D}) &= \mathcal{W}(\hat{\mathbf{T}}(\beta, \theta, \psi,  \mathbf{D}), J(\beta), \theta, \hat{W})  \\
    \hat{\mathbf{T}}(\beta, \theta, \psi,  \mathbf{D}) &= \mathcal{S}(\mathbf{T}(\beta, \theta, \psi)) + \mathbf{D}  ,
\end{split}
\end{align}
where $\mathcal{S}: \mathbb{R}^{N \times 3}  \rightarrow \mathbb{R}^{N_s \times 3}$ is the mesh subdivision operation, $ \mathbf{D} \in \mathbb{R}^{N_s\times 3}, \hat{W}\in \mathbb{R}^{N_s\times J}$ and $N_s$ are the vertex displacement, skinning weights and vertices number of the subdivided body, respectively.
Note that besides the displacement $\mathbf{D}$, the parameters $\beta, \theta, \psi$ are also learnable.
This helps to generate various kinds of characters with various shapes, such as human-like characters and anime characters with large deformed body shapes, like exaggerated proportions, elongated limbs, large eyes, and \etc.

\paragraph{Partial Mesh Subdivision.}\label{sec:mesh_subdivision}
The vertices on the surface of the SMPL-X body are irregularly distributed, i.e., around 4,000 vertices are located on the head with the remaining 6,000 on the body.
The sparsity of vertices on the body surface results in less detailed deformations there. 
Simply increasing the mesh density by subdividing the whole body mesh leads to noisy results, especially, in the face area during geometry optimization.
To address this issue, we employ an adaptive upsampling technique on the triangles and interpolate their skinning weights within areas of low mesh density, such as the body region and the back of the head. 
This process yields a more refined mesh with uniformly distributed vertices and smoother skinning weights.
%
%


\subsection{Consistent Geometry and Texture Learning} 
\label{sec:geo_tex_learning}
To generate animatable characters, we need to ensure the consistency between geometry and the texture.
Therefore, we propose to blend the SDS loss of the rendered normal and RGB images to achieve a well-aligned geometry and texture.
Given a mesh $\hat{\mathbf{M}}$ parameterized by $\mathbf{D}, \beta$ and $\psi$ and albedo $\Psi$, we render its normal image $\mathcal{N}$ and colored image $\mathcal{I}$ using a differentiable render~\cite{Laine2020diffrast}, denoted as $\mathcal{R}$:
\begin{equation} 
    \mathcal{N} =  \mathcal{R} (\hat{\mathbf{M}},\pi), \quad \mathcal{I} =  \mathcal{R} (\Psi, \hat{\mathbf{M}},\pi)
\end{equation}
where $\pi$ are the camera parameters.
In each iteration, the camera is randomly positioned in one of two perspectives: a full-body view or a zoom-in head view. 
The head zoom-in allows us to reconstruct a detailed face region.

\qheading{Texture SDS Objective} 
Given a text prompt, the texture generation is guided by a pretrained Stable Diffusion (SD) model~\cite{rombach2022high}, denoted as $\phi$, which measures the similarity between the rendered image and the provided text prompt within the added and predicted noise space:
\begin{equation}
\label{eq:tex}
    \nabla_{\Psi} \mathcal{L}_{\mathrm{tex}}(\phi, {\mathcal{I}}) = \mathbb{E}_{t,\epsilon} \left[ w(t)(\hat{\epsilon}_{\phi}(z_t^\mathcal{I}; y, t) - \epsilon) \frac{\partial  \mathcal{I}}{\partial \Psi} \frac{\partial  z^\mathcal{I}}{\partial \mathcal{I}}  \right] ,
\end{equation}  
where $z^\mathcal{I}$ is the latent feature of $ \mathcal{I}$, encoded by image encoder (SD), $\hat{\epsilon}_{\phi}(z_t^\mathcal{I}; y, t)$ is the predicted noise given text embedding $y$ and noise level $t$, $\epsilon$ is the pre-computed noise.

\qheading{Geometry Consistency SDS Objective} 
Similarly, rendered normal images can be used for the diffusion model as shape encoding to facilitate the geometry synthesis.
However, this approach may encounter challenges in ensuring perfect consistency between geometry and texture. 
To address this issue, we compute the SDS loss on the interpolation between normal and color image latents.
\begin{equation}
\label{eq:consistency}
    \nabla_{\gamma } \mathcal{L}_{\mathrm{c}}(\phi, x) = \mathbb{E}_{t,\epsilon} \left[ w(t)(\hat{\epsilon}_{\phi}(\tilde{z_t}; y, t) - \epsilon) \frac{\partial  \mathcal{N}}{\partial \gamma } \frac{\partial  z}{\partial \mathcal{N}}  \right] ,
\end{equation}   
where $\gamma=\{\beta, \psi, \mathbf{D}\}$ are the geometry related parameters, $\tilde{z}=\alpha z^{\mathcal{I}} + (1-\alpha)z^{\mathcal{N}}$ denotes the resulting interpolated latent code, while $z^{\mathcal{I}}$ and $z^{\mathcal{N}}$
represent the latent codes corresponding to the RGB and  normal image, respectively.
%
%

\qheading{Overall Optimization Objective} 
The learning objectives can be formulated as a combination of the texture SDS objective $\mathcal{L}_{\mathrm{tex}}$ and the geometry consistency loss $\mathcal{L}_{\mathrm{c}}$, where $\lambda_{\mathrm{tex}}$ and $\lambda_\mathrm{c}$ are the corresponding loss weights:
\begin{equation}
\label{eq:loss}
    \mathcal{L} = \lambda_{\mathrm{tex}} \mathcal{L}_{\mathrm{tex}} + \lambda_{\mathrm{c}} \mathcal{L}_{\mathrm{c}} ,
\end{equation}
Based on~\cref{eq:loss}, the geometry and texture are optimized jointly.
We employ a progressive optimization strategy for the rendered color image $\mathcal{I}$ in the \Eref{eq:tex}. Initially, this image is generated at a low resolution ($32 \times 32$), which gradually increases during the optimization process, ultimately reaching $512 \times 512$ resolution. In contrast, both the rendered normal image $\mathcal{N}$ and color image $\mathcal{I}$ in the \Eref{eq:consistency} remain $512 \times 512$ resolution throughout the entire procedure. 
Additionally, we detach the gradients of $z^{\mathcal{I}}$ in \cref{eq:consistency}, allowing only geometric updates, while optimizing textures using the texture SDS loss. This approach ensures both texture-text consistency and geometry-texture alignment, preventing misalignment that could result in unrealistic animation. 
 

\subsection{Training with Animations}
\label{sec:multi_expression_training}
To ensure plausible animations, particularly for the face region, it is essential to maintain semantic correspondence with the SMPL-X model.
However, during optimization, certain parts may undergo changes and not align perfectly with the original ones (\eg the mouth may be mapped to the chin area or become distorted).
If not addressed, animated results will have severe artifacts as the wrong parts will be deformed with the SMPL-X  model.
To tackle this problem, we optimize the avatar using various animations (see Fig.~\ref{fig:pipeline}). 
In particular, we find that using different jaw poses during training helps produce well aligned faces.
We found that animating the SMPL-X expression parameters, made little visible difference.
We suspect that these would become relevant with an even higher-resolution face mesh.
%
%
Specifically, during optimization, we randomly sample one jaw pose in each iteration from an expression gallery $\Theta$, i.e., a motion sequence from TalkSHOW \cite{yi2022talkshow}. 
The final optimization process minimizes the following objective:
\begin{equation}
    \min_{\beta, \psi, \mathbf{D}, \Psi} \mathbb{E}_{\theta \in \Theta} \left[ \mathcal{L}(\phi, x(\beta, \theta, \psi, \mathbf{D}, \Psi)) \right] .
\end{equation}

\begin{figure*}[t]
\centering
\includegraphics[width=\linewidth]{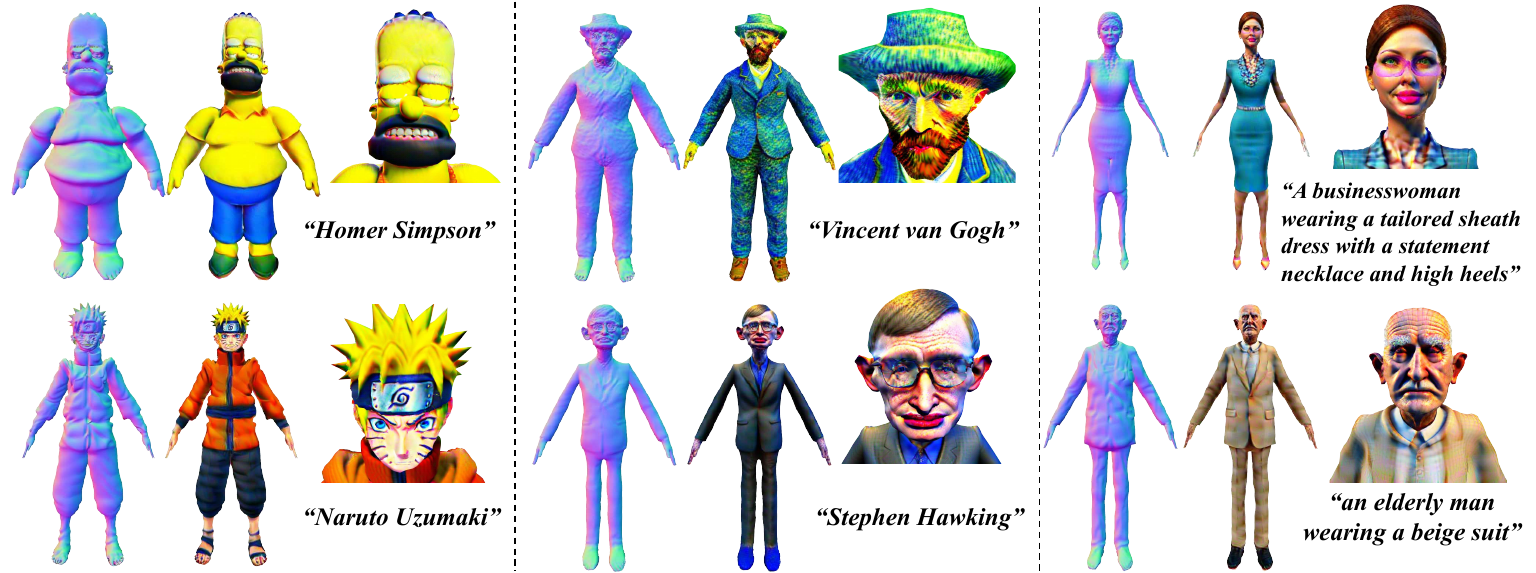}  
\caption{\textbf{Diverse Range of Avatar Generation}. \modelname has the ability to generate a broad spectrum of characters, which includes iconic figures, celebrities, and customized avatars based on textual descriptions. }
\label{fig:avatar}
\end{figure*}

\section{Experiments}
\label{sec:experiments}
We first demonstrate our expressive, holistic, animation of the avatars, then evaluate their quality, and the consistency between texture and geometry.
%
Finally, ablation studies are conducted to analyze the effectiveness of each component. 

\begin{figure*}[t]
\centering 
\includegraphics[ 
width=\linewidth]{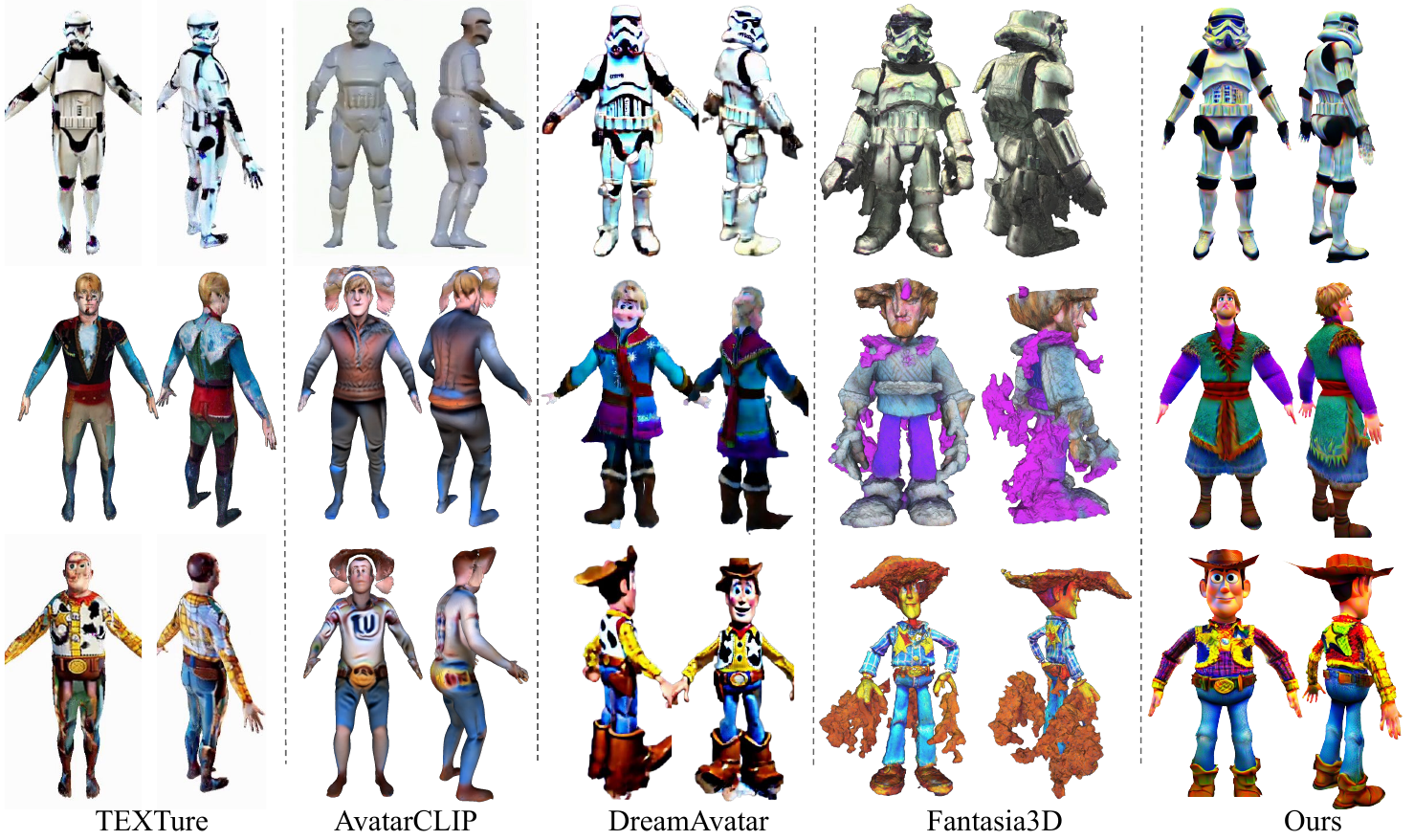}
\caption{\textbf{Qualitative comparison}. The prompts (top $\rightarrow$ down) are \textit{``Stormtrooper'', ``Woody in Toy Story'', ``Kristoff in Frozen''}. Compared with baselines using: A) body mesh w/o displacement clothing layer (TEXTure~\cite{richardson2023texture}, AvatarCLIP~\cite{hong2022avatarclip}), B) Neural fields (DreamAvatar~\cite{cao2023dreamavatar}), C) \textsc{DMTet} (Fantasia3D~\cite{chen2023fantasia3d}). \modelname generates more high-quality characters in terms of both geometry and texture.}
\label{fig:compare}
\end{figure*}

\subsection{Expressive Holistic Body Animation} 
\label{sec:animation}
One crucial feature that distinguishes our method from others is that \modelname enables natural full-body animations over the face, body and hands.
Figure \ref{fig:animation} illustrates the animation of characters generated by \modelname using  only with text as input.
In the first case, we convert text to audio \cite{tts} and then use TalkSHOW \cite{yi2022talkshow} create expressive SMPL-X animations of the upper body, face and hands.
In the second case we use priorMDM \cite{shafir2023human} to convert text into SMPL \cite{SMPL:2015} animations, which we convert to SMPL-X \cite{SMPL-X:2019}.
%
Thanks semantic correspondence with SMPL-X, the characters are easily animated with natural movements of the fully body and face.
%
This consistency with SMPL-X means that avatars generated by \modelname can be animated using any of the recent text to animation methods that output SMPL-X.


\subsection{Diverse Range of Avatars}
%
%
As shown in \Fref{fig:avatar}, \modelname produces a wide variety of 3D avatars characterized by their high-quality geometry and realistic textures. These avatars contain fictional characters from animated films, real-life celebrities, and custom-made characters based on prompts generated by ChatGPT.
%
%
This capability opens up numerous real-world applications, enabling users to effortlessly generate avatars with a wide range of shapes, appearances, and clothing styles.

\subsection{Qualitative Comparison}
\label{sec:qualitative_comparison}
We compare our method with existing approaches on the task of text-to-3D human avatar generation.
We consider four state-of-the-art methods for full body avatar generation, where the original implementation is available: TEXTure~\cite{richardson2023texture}, AvatarCLIP~\cite{hong2022avatarclip}, DreamAvatar~\cite{cao2023dreamavatar} and Fantasia3D~\cite{chen2023fantasia3d}. 
For head-only avatar generation, we compare our approach with DreamFace~\cite{zhang2023dreamface} and HeadSculpt~\cite{han2023headsculpt}.

\begin{figure}[t]
\centering
\vspace{-1.5 em}
\includegraphics[width=\linewidth]{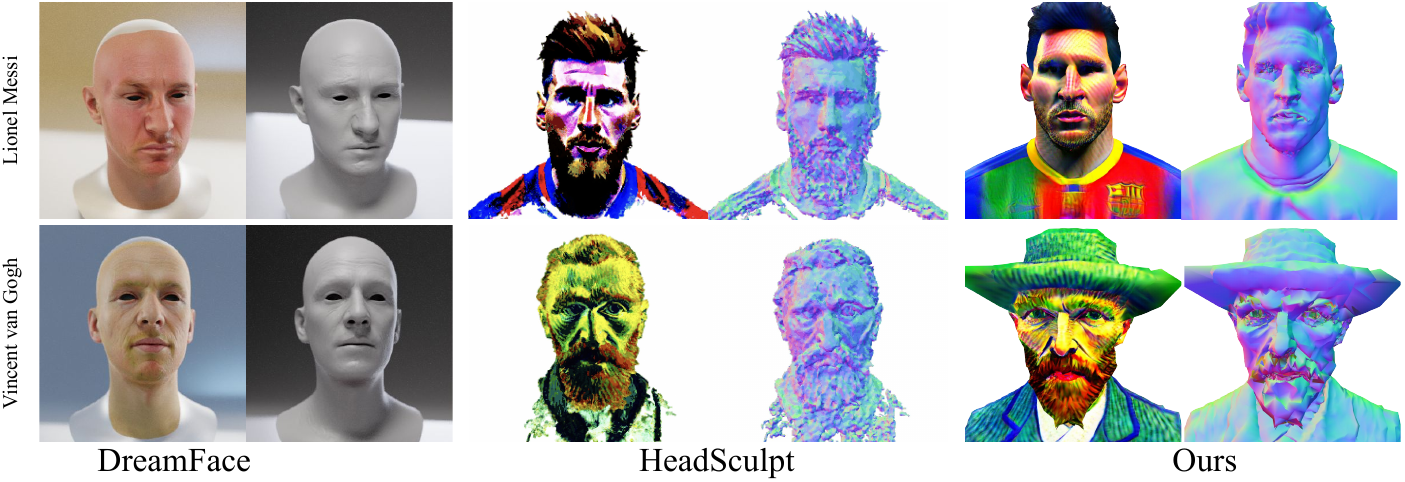}  
\caption{\textbf{Comparison of head generator}. 
While DreamFace excels in generating CG-compatible facial assets, it struggles with shapes that deviate significantly from the norm, such as accessories like hats. HeadSculpt often produces noisy artifacts in its output. In contrast, \modelname generates a broader range of detailed shapes and appearances with greater fidelity.
}
\label{fig:compare_face}
\end{figure}

\begin{table}[t]
\centering
\setlength{\tabcolsep}{2.5pt}
\resizebox{0.8\linewidth}{!}{
\begin{tabular}{c|ccc}
Preference (\%, $\uparrow$)   &  AvatarCLIP  & DreamAvatar   \\ \shline
Geometry Quality &  94.45   &  87.77  \\
Texture Quality  & 94.74  & 82.67 \\
Consistency with Input Prompt  & 95.00 & 81.52\\

\end{tabular}
}
\caption{\textbf{User Study}. User preference results indicates that \modelname significantly outperforms other baselines in terms of geometry, texture, and consistency with the input prompt. Its superior performance is evident across all three key aspects.}
\vspace{-1.5 em}
\label{table:user-study}
\end{table}

\medskip
\noindent \textbf{Full Body Avatar.}
%
Figure \ref{fig:compare} provides a qualitative comparison of  avatars generated by \cite{richardson2023texture, hong2022avatarclip, cao2023dreamavatar} and by \modelname. 
In comparison to the baselines, \modelname  generates avatars with considerably more realistic textures.
Also \modelname produces a wide range of 3D body shapes (cf.~TEXTure), without geometric artifacts (cf.~AvatarCLIP, DreamAvatar and Fantasia3D), and with a semantically correct texture that is consistent with the geometry.
%


\medskip
\noindent \textbf{Head Avatar.} 
Furthermore, \modelname generates high-quality head avatars as shown in Fig.~\ref{fig:compare_face}.
We compare with DreamFace~\cite{zhang2023dreamface} and HeadSculpt~\cite{han2023headsculpt}, a shape sculpting method specifically designed for head avatar generation.
%
Note that \modelname creates visually appealing head avatars with consistent and well-aligned geometry as well as high-fidelity textures.
However, others have different limitations.
%
DreamFace~\cite{zhang2023dreamface} avatars can look realistic but are strongly biased towards natural head shapes and cannot capture more varied facial details like mustaches or cartoon shapes.
Like our results, the head avatars can be animated.
%
Meanwhile, HeadScuplt~\cite{han2023headsculpt} generates noisy geometry and texture, making the output less useful for downstream tasks like animation.
%
%


\subsection{Quantitative Evaluation.} 
To quantitatively evaluate \modelname, we conducted a A/B user study with 17 CV students.
We used a Google Survey Form to assess the (1) geometry quality, (2) texture quality, and (3) consistency with input prompts.
We used ChatGPT to automatically generate a large set of character descriptions, including celebrities, characters in movies and anime, and general occupation character descriptions, select 27 of these at random, and generate the corresponding avatars;
see Sup.~Mat.~for details.
%
%
In A/B tests, the participants were asked to select the preferred reconstruction from randomly selected videos from the baselines (AvatarCLIP and DreamAvatar) and our method (see \tref{table:user-study}).
%
The results show that our proposed method achieves considerable higher preference over the baseline methods over all three metrics.
%




\begin{figure}[t!]
\centering
\vspace{-1.5 em}
\includegraphics[width=\linewidth]{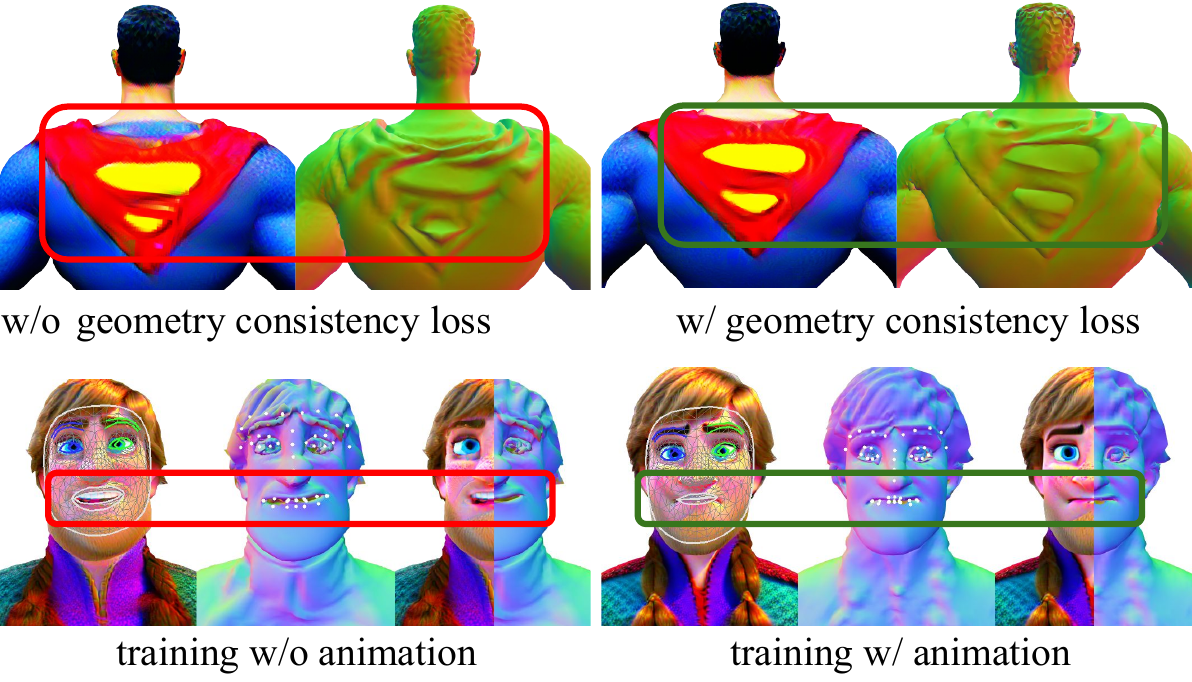}   
\vspace{-1.5 em} 
\caption{\textbf{Ablation study on 1) geometry consistency loss, and 2) training with animation}. The geometry consistency loss generates better well-aligned geometry and texture, while training with animation helps remain the semantic correspondences with the original SMPL-X, especially at the mouth region. 
}
\label{fig:ablation}
\vspace{-1.0 em} 
\end{figure}

\subsection{Ablation Study}
We conducted ablation studies to evaluate the effects of the geometry consistency loss and the optimization with animations in our method. 
The results shown in \fref{fig:ablation} demonstrate the effectiveness of these components. 
The consistency loss improves the alignment between the geometry and texture on the backside of the ``Superman'', while training with animations improves the face geometry by enforcing the semantic correspondence with SMPL-X, particularly at mouth region. 
These advancements enable us to effortlessly animate our high-resolution avatars, leveraging the pose and expression space of the SMPL-X model.

 



\begin{figure*}[h!]
\centering
\vspace{0.5 em}
\includegraphics[width=\linewidth]{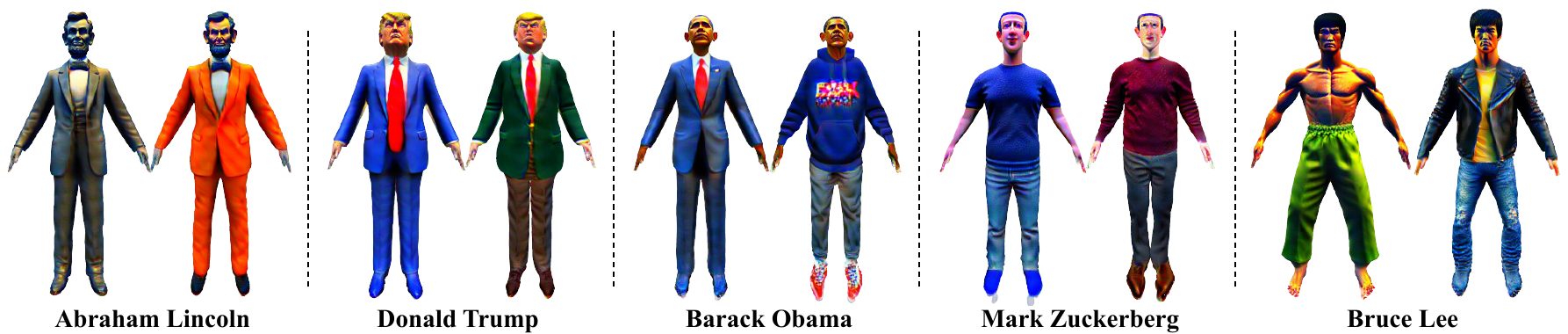}  
\caption{\textbf{Virtual Try-on}. We demonstrate five individuals: Abraham Lincoln, Donald Trump, Barack Obama, Mark Zuckerberg, and Bruce Li, each with two different outfits. The first one represents their typical dressing style, personalized according to their known preferences with their name as the only input. 
The other one is imagined by ChatGPT, complete with detailed descriptions of the attire.}
\label{fig:tryon}
\end{figure*} 

\section{Editing Applications}  
\modelname facilitates several applications, such as virtual try-on, text-guided texture editing, and local geometry transferring. 

\begin{figure}[ht!]
\centering
\includegraphics[width=\linewidth]{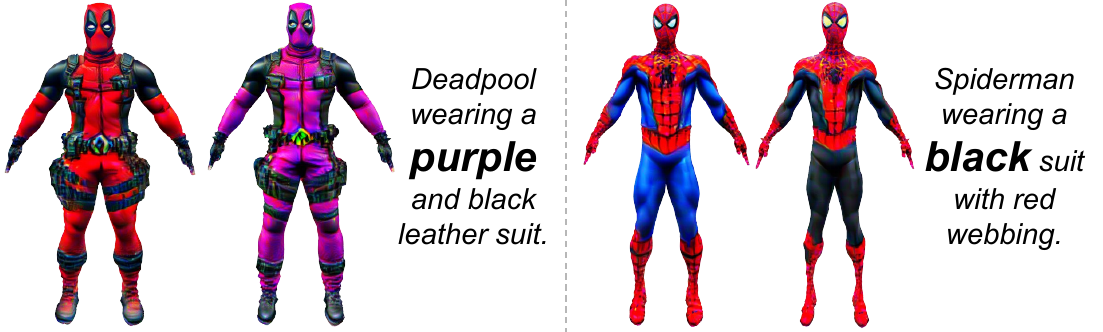}  
\caption{\textbf{Text-guided texture editing}. \modelname possesses the ability to modify the color of clothing via changing texts. }
\label{fig:color_editing}
\end{figure}

\medskip
\qheading{Virtual Try-on} 
\modelname can be used for virtual try-on, i.e., we can ask ChatGPT \cite{chatgpt} to design fashion outfits for a specific person as depicted in \Fref{fig:tryon}. 
The visual results indicate that our method can generate avatars with text-guided personalized textures while preserving the identity. 

\qheading{Texture Editing}
Figure \ref{fig:color_editing} shows examples of modifying outfit textures by changing the input text.
This is particularly valuable for film or game character design, allowing easy alterations such as changing the color palette of a character.
%
Designers can quickly visualize their desired aesthetic appeal and bring their creative vision to life.
%

\begin{figure}[ht!]
\centering
\includegraphics[width=\linewidth]{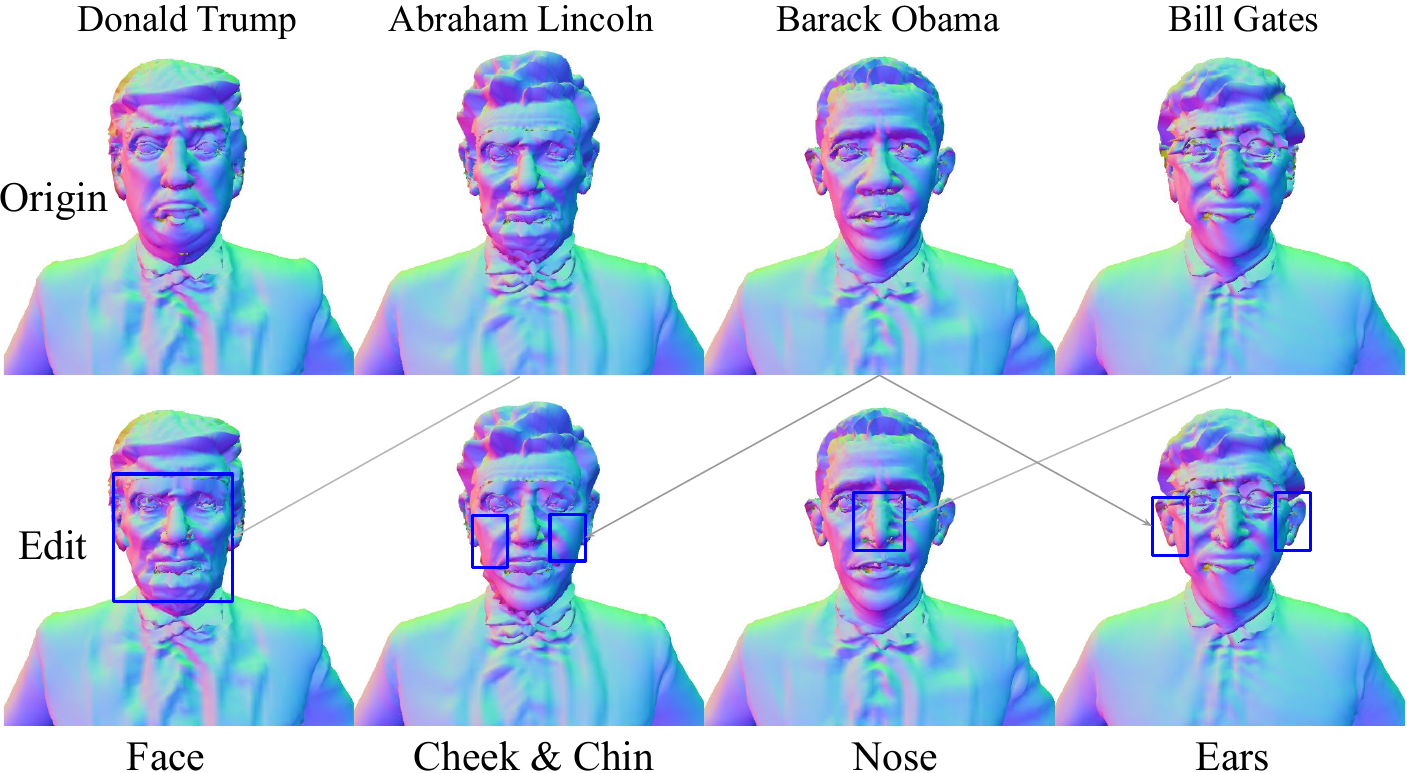}  
\caption{\textbf{Local Shape Editing}. We demonstrate an example of face swapping across four different celebrities.}
\label{fig:face_swap}
\end{figure}  

\qheading{Local Shape Editing}
\label{sec:application_face_swap} 
Thanks to the body-part segments of SMPL-X, our method supports direct local body and face swapping between two avatars without any additional effort. \Fref{fig:face_swap} gives an example of face editing on four individuals. This is also applicable to body or clothing transferring. In addition to geometry or texture transferring, \modelname can also be utilized for local shape sculpting through user-friendly prompts as input. This feature is particularly helpful for artists in designing customized avatars.

\section{Discussion}
\label{sec:discussion}

While \modelname shows promising results, it still has several limitations. Additionally, further investigation is needed to assess any potential negative social impact.

\medskip
\qheading{Limitations \& Future works}
One aspect that requires improvement is the \textit{relighting capabilities} in different environments, e.g. generated indoor rooms from MIME~\cite{yi2023mime}, thus enabling photo-realistic rendering with human-scene interactions. This can benefit from using BRDF, by separating the texture into separate components (\ie, material, albedo, and lighting) like Fantasia3D~\cite{chen2023fantasia3d}. 
%
Furthermore, \modelname can generate avatars with diverse body shapes, some of which may deviate largely from the base SMPL-X model. In such cases, using the original skinning weights may lead to unrealistic animations. Therefore, exploring the joint learning of \textit{adaptable skinning weights} specifically tailored to text input could be a promising direction. 
Textual descriptions alone may not fully capture the nuanced and intricate aspects of a character's appearance.
Combining existing controllable text-to-image models \cite{mou2023t2i,zhang2023adding,oft2023} can be beneficial to provide more detailed control over a character's face or clothing. And the \textit{compositional generation} of separate haircut~\cite{sklyarova2023neural}, accessories~\cite{gao2022dart}, and decoupled outfits~\cite{Feng2022scarf} could also be a valuable exploration direction.

\medskip
\qheading{Social Impact}
As the technique progresses, it raises concerns about deep-fake and intellectual property (IP) when we generate iconic characters. Regulations should be established to address these issues alongside the benefits in the entertainment industry. Additionally, it is crucial to prioritize gender and cultural diversity. For instance, if the term ``police officer'' consistently generates a male instead of considering both genders, it implies potential gender bias. Ensuring inclusivity and avoiding stereotypes are essential in mitigating any adverse social impact.

\section{Conclusion}
\label{sec:conclusion}

We introduce \modelname, a simple yet effective method for generating high-quality and animatable 3D textured avatars solely from text input. These avatars cover a wide range of individuals, including celebrities and customized characters. They seamlessly integrate into existing CG pipelines, catering to various industries like fashion and entertainment. The key contributions include: 1) utilizing a subdivided version of SMPL-X with learned displacement layer and UV texture, 2) employing hierarchical optimization with adaptive focal lengths, 3) enforcing geometry-texture alignment through geometric consistency loss, and 4) training with animation to keep semantic correspondence with SMPL-X. We validate these components through ablation studies and demonstrate the superiority of \modelname over other SOTAs with both qualitative and quantitative results.




\mheading{Acknowledgments}
Thanks Zhen Liu and Weiyang Liu for their fruitful discussion, Haofan Wang and Xu Tang for their technical support, and Benjamin Pelkofer for IT support.
Hongwei Yi is supported in part by the German Federal Ministry of Education and Research (BMBF): Tübingen AI Center, FKZ: 01IS18039B. 
Yuliang Xiu is funded by the European Union’s Horizon $2020$ research and innovation programme under the Marie Skłodowska-Curie grant agreement No.$860768$ (\href{https://www.clipe-itn.eu}{CLIPE}).
Jiaxiang Tang is supported by National Natural Science Foundation of China (Grant Nos: 61632003, 61375022, 61403005).
Yangyi Huang is supported by the National Nature Science Foundation of China (Grant Nos: 62273302, 62036009, 61936006).

\mheading{Disclosure}
MJB has received research gift funds from Adobe, Intel, Nvidia, Meta/Facebook, and Amazon. MJB has financial interests in Amazon, Datagen Technologies, and Meshcapade GmbH.  While MJB is a consultant for Meshcapade, his research in this project was performed solely at, and funded solely by, the Max Planck Society.


{
    \small
    \bibliographystyle{ieeenat_fullname}
    \bibliography{main}

\begin{thebibliography}{80}
\providecommand{\natexlab}[1]{#1}
\providecommand{\url}[1]{\texttt{#1}}
\expandafter\ifx\csname urlstyle\endcsname\relax
  \providecommand{\doi}[1]{doi: #1}\else
  \providecommand{\doi}{doi: \begingroup \urlstyle{rm}\Url}\fi

\bibitem[Alldieck et~al.(2021)Alldieck, Xu, and
  Sminchisescu]{alldieck2021imghum}
Thiemo Alldieck, Hongyi Xu, and Cristian Sminchisescu.
\newblock imghum: Implicit generative models of 3d human shape and articulated
  pose.
\newblock In \emph{{International Conference on Computer Vision ({ICCV})}},
  pages 5461--5470, 2021.

\bibitem[Aneja et~al.(2023)Aneja, Thies, Dai, and
  Nie{\ss}ner]{aneja2023clipface}
Shivangi Aneja, Justus Thies, Angela Dai, and Matthias Nie{\ss}ner.
\newblock Clipface: Text-guided editing of textured 3d morphable models.
\newblock In \emph{SIGGRAPH '23 Conference Proceedings}, 2023.

\bibitem[Bickel et~al.(2007)Bickel, Botsch, Angst, Matusik, Otaduy, Pfister,
  and Gross]{bickel2007multi}
Bernd Bickel, Mario Botsch, Roland Angst, Wojciech Matusik, Miguel Otaduy,
  Hanspeter Pfister, and Markus Gross.
\newblock Multi-scale capture of facial geometry and motion.
\newblock \emph{{Transactions on Graphics (TOG)}}, 26\penalty0 (3):\penalty0
  33--es, 2007.

\bibitem[Brown et~al.(2020)Brown, Mann, Ryder, Subbiah, Kaplan, Dhariwal,
  Neelakantan, Shyam, Sastry, Askell, et~al.]{brown2020language}
Tom Brown, Benjamin Mann, Nick Ryder, Melanie Subbiah, Jared~D Kaplan, Prafulla
  Dhariwal, Arvind Neelakantan, Pranav Shyam, Girish Sastry, Amanda Askell,
  et~al.
\newblock Language models are few-shot learners.
\newblock In \emph{{Conference on Neural Information Processing Systems
  (NeurIPS)}}, pages 1877--1901, 2020.

\bibitem[Cao et~al.(2023)Cao, Cao, Han, Shan, and Wong]{cao2023dreamavatar}
Yukang Cao, Yan-Pei Cao, Kai Han, Ying Shan, and Kwan-Yee~K Wong.
\newblock {DreamAvatar: Text-and-Shape Guided 3D Human Avatar Generation via
  Diffusion Models}.
\newblock \emph{arXiv preprint:2304.00916}, 2023.

\bibitem[chatgpt()]{chatgpt}
chatgpt.
\newblock \url{https://chat.openai.com/}, 2022.

\bibitem[Chen et~al.(2023)Chen, Chen, Jiao, and Jia]{chen2023fantasia3d}
Rui Chen, Yongwei Chen, Ningxin Jiao, and Kui Jia.
\newblock {Fantasia3D: Disentangling Geometry and Appearance for High-quality
  Text-to-3D Content Creation}.
\newblock In \emph{{International Conference on Computer Vision ({ICCV})}},
  2023.

\bibitem[Chen et~al.(2022)Chen, Chen, Lei, Zhang, and Jia]{chen2022tango}
Yongwei Chen, Rui Chen, Jiabao Lei, Yabin Zhang, and Kui Jia.
\newblock {TANGO: Text-driven Photorealistic and Robust 3D Stylization via
  Lighting Decomposition}.
\newblock In \emph{{Conference on Neural Information Processing Systems
  (NeurIPS)}}, 2022.

\bibitem[dalle2()]{dalle2}
dalle2.
\newblock \url{https://openai.com/dall-e-2}, 2022.

\bibitem[Feng et~al.(2022)Feng, Yang, Pollefeys, Black, and
  Bolkart]{Feng2022scarf}
Yao Feng, Jinlong Yang, Marc Pollefeys, Michael~J. Black, and Timo Bolkart.
\newblock {Capturing and Animation of Body and Clothing from Monocular Video}.
\newblock In \emph{SIGGRAPH Asia 2022 Conference Papers}, 2022.

\bibitem[Fu et~al.(2022)Fu, Li, Jiang, Lin, Qian, Loy, Wu, and
  Liu]{fu2022styleganhuman}
Jianglin Fu, Shikai Li, Yuming Jiang, Kwan-Yee Lin, Chen Qian, Chen-Change Loy,
  Wayne Wu, and Ziwei Liu.
\newblock Stylegan-human: A data-centric odyssey of human generation.
\newblock \emph{{European Conference on Computer Vision (ECCV)}}, 2022.

\bibitem[Gao et~al.(2022)Gao, Xiu, Li, Yang, Wang, Zhang, Zhang, Lu, and
  Tan]{gao2022dart}
Daiheng Gao, Yuliang Xiu, Kailin Li, Lixin Yang, Feng Wang, Peng Zhang, Bang
  Zhang, Cewu Lu, and Ping Tan.
\newblock {DART: Articulated Hand Model with Diverse Accessories and Rich
  Textures}.
\newblock In \emph{Thirty-sixth Conference on Neural Information Processing
  Systems Datasets and Benchmarks Track}, 2022.

\bibitem[Gao et~al.(2023{\natexlab{a}})Gao, Aigerman, Thibault, Kim, and
  Hanocka]{Gao_2023_SIGGRAPH}
William Gao, Noam Aigerman, Groueix Thibault, Vladimir Kim, and Rana Hanocka.
\newblock Textdeformer: Geometry manipulation using text guidance.
\newblock In \emph{ACM Transactions on Graphics (SIGGRAPH)},
  2023{\natexlab{a}}.

\bibitem[Gao et~al.(2023{\natexlab{b}})Gao, Aigerman, Thibault, Kim, and
  Hanocka]{TextDeformer}
William Gao, Noam Aigerman, Groueix Thibault, Vladimir Kim, and Rana Hanocka.
\newblock Textdeformer: Geometry manipulation using text guidance.
\newblock In \emph{ACM Transactions on Graphics (SIGGRAPH)},
  2023{\natexlab{b}}.

\bibitem[Habermann et~al.(2020)Habermann, Xu, Zollhoefer, Pons-Moll, and
  Theobalt]{deepcap}
Marc Habermann, Weipeng Xu, Michael Zollhoefer, Gerard Pons-Moll, and Christian
  Theobalt.
\newblock Deepcap: Monocular human performance capture using weak supervision.
\newblock In \emph{{Computer Vision and Pattern Recognition (CVPR)}}. {IEEE},
  2020.

\bibitem[Han et~al.(2023)Han, Cao, Han, Zhu, Deng, Song, Xiang, and
  Wong]{han2023headsculpt}
Xiao Han, Yukang Cao, Kai Han, Xiatian Zhu, Jiankang Deng, Yi-Zhe Song, Tao
  Xiang, and Kwan-Yee~K. Wong.
\newblock Headsculpt: Crafting 3d head avatars with text.
\newblock \emph{arXiv preprint arXiv:2306.03038}, 2023.

\bibitem[Ho et~al.(2020)Ho, Jain, and Abbeel]{ho2020denoising}
Jonathan Ho, Ajay Jain, and Pieter Abbeel.
\newblock Denoising diffusion probabilistic models.
\newblock In \emph{{Conference on Neural Information Processing Systems
  (NeurIPS)}}, pages 6840--6851, 2020.

\bibitem[Hong et~al.(2022)Hong, Zhang, Pan, Cai, Yang, and
  Liu]{hong2022avatarclip}
Fangzhou Hong, Mingyuan Zhang, Liang Pan, Zhongang Cai, Lei Yang, and Ziwei
  Liu.
\newblock {AvatarCLIP: Zero-Shot Text-Driven Generation and Animation of 3D
  Avatars}.
\newblock \emph{{Transactions on Graphics (TOG)}}, 41\penalty0 (4):\penalty0
  1--19, 2022.

\bibitem[Huang et~al.(2023{\natexlab{a}})Huang, Wang, Zeng, Cao, Qi, Shi, Zha,
  and Zhang]{huang2023dreamwaltz}
Yukun Huang, Jianan Wang, Ailing Zeng, He Cao, Xianbiao Qi, Yukai Shi,
  Zheng-Jun Zha, and Lei Zhang.
\newblock Dreamwaltz: Make a scene with complex 3d animatable avatars.
\newblock \emph{arXiv preprint arXiv:2305.12529}, 2023{\natexlab{a}}.

\bibitem[Huang et~al.(2023{\natexlab{b}})Huang, Yi, Liu, Wang, Wu, Wang, Lin,
  Zhang, and Cai]{huang2022elicit}
Yangyi Huang, Hongwei Yi, Weiyang Liu, Haofan Wang, Boxi Wu, Wenxiao Wang,
  Binbin Lin, Debing Zhang, and Deng Cai.
\newblock One-shot implicit animatable avatars with model-based priors.
\newblock In \emph{{International Conference on Computer Vision ({ICCV})}},
  2023{\natexlab{b}}.

\bibitem[Jain et~al.(2022)Jain, Mildenhall, Barron, Abbeel, and
  Poole]{jain2021dreamfields}
Ajay Jain, Ben Mildenhall, Jonathan~T. Barron, Pieter Abbeel, and Ben Poole.
\newblock Zero-shot text-guided object generation with dream fields.
\newblock In \emph{{Computer Vision and Pattern Recognition (CVPR)}}, 2022.

\bibitem[Joo et~al.(2015)Joo, Liu, Tan, Gui, Nabbe, Matthews, Kanade, Nobuhara,
  and Sheikh]{joo2015panoptic}
Hanbyul Joo, Hao Liu, Lei Tan, Lin Gui, Bart Nabbe, Iain Matthews, Takeo
  Kanade, Shohei Nobuhara, and Yaser Sheikh.
\newblock Panoptic studio: A massively multiview system for social motion
  capture.
\newblock In \emph{{Computer Vision and Pattern Recognition (CVPR)}}, pages
  3334--3342, 2015.

\bibitem[Karras et~al.(2020)Karras, Laine, Aittala, Hellsten, Lehtinen, and
  Aila]{Karras2019stylegan2}
Tero Karras, Samuli Laine, Miika Aittala, Janne Hellsten, Jaakko Lehtinen, and
  Timo Aila.
\newblock Analyzing and improving the image quality of {StyleGAN}.
\newblock In \emph{{Computer Vision and Pattern Recognition (CVPR)}}, 2020.

\bibitem[Kim et~al.(2023)Kim, Kwon, Lee, Lee, Han, Kim, and Joo]{kim2023chupa}
Byungjun Kim, Patrick Kwon, Kwangho Lee, Myunggi Lee, Sookwan Han, Daesik Kim,
  and Hanbyul Joo.
\newblock {Chupa: Carving 3D Clothed Humans from Skinned Shape Priors using 2D
  Diffusion Probabilistic Models}.
\newblock In \emph{{International Conference on Computer Vision ({ICCV})}},
  2023.

\bibitem[Kolotouros et~al.(2023)Kolotouros, Alldieck, Zanfir, Bazavan, Fieraru,
  and Sminchisescu]{kolotouros2023dreamhuman}
Nikos Kolotouros, Thiemo Alldieck, Andrei Zanfir, Eduard~Gabriel Bazavan, Mihai
  Fieraru, and Cristian Sminchisescu.
\newblock Dreamhuman: Animatable 3d avatars from text.
\newblock \emph{arXiv preprint:2306.09329}, 2023.

\bibitem[Laine et~al.(2020)Laine, Hellsten, Karras, Seol, Lehtinen, and
  Aila]{Laine2020diffrast}
Samuli Laine, Janne Hellsten, Tero Karras, Yeongho Seol, Jaakko Lehtinen, and
  Timo Aila.
\newblock Modular primitives for high-performance differentiable rendering.
\newblock \emph{{Transactions on Graphics (TOG)}}, 39\penalty0 (6), 2020.

\bibitem[Li et~al.(2020{\natexlab{a}})Li, Olszewski, Xiu, Saito, Huang, and
  Li]{li2020monoportRTL}
Ruilong Li, Kyle Olszewski, Yuliang Xiu, Shunsuke Saito, Zeng Huang, and Hao
  Li.
\newblock Volumetric human teleportation.
\newblock In \emph{ACM SIGGRAPH 2020 Real-Time Live}, 2020{\natexlab{a}}.

\bibitem[Li et~al.(2020{\natexlab{b}})Li, Xiu, Saito, Huang, Olszewski, and
  Li]{li2020monoport}
Ruilong Li, Yuliang Xiu, Shunsuke Saito, Zeng Huang, Kyle Olszewski, and Hao
  Li.
\newblock {Monocular real-time volumetric performance capture}.
\newblock In \emph{{European Conference on Computer Vision (ECCV)}}, pages
  49--67. Springer, 2020{\natexlab{b}}.

\bibitem[Liao et~al.(2023)Liao, Zhang, Xiu, Yi, Liu, Qi, Zhang, Wang, Zhu, and
  Lei]{liao2023car}
Tingting Liao, Xiaomei Zhang, Yuliang Xiu, Hongwei Yi, Xudong Liu, Guo-Jun Qi,
  Yong Zhang, Xuan Wang, Xiangyu Zhu, and Zhen Lei.
\newblock {High-Fidelity Clothed Avatar Reconstruction from a Single Image}.
\newblock In \emph{{Computer Vision and Pattern Recognition (CVPR)}}, 2023.

\bibitem[Lin et~al.(2023)Lin, Gao, Tang, Takikawa, Zeng, Huang, Kreis, Fidler,
  Liu, and Lin]{lin2022magic3d}
Chen-Hsuan Lin, Jun Gao, Luming Tang, Towaki Takikawa, Xiaohui Zeng, Xun Huang,
  Karsten Kreis, Sanja Fidler, Ming-Yu Liu, and Tsung-Yi Lin.
\newblock {Magic3D: High-Resolution Text-to-3D Content Creation}.
\newblock In \emph{{Computer Vision and Pattern Recognition (CVPR)}}, 2023.

\bibitem[Loper et~al.(2015)Loper, Mahmood, Romero, Pons-Moll, and
  Black]{SMPL:2015}
Matthew Loper, Naureen Mahmood, Javier Romero, Gerard Pons-Moll, and Michael~J.
  Black.
\newblock {SMPL}: A skinned multi-person linear model.
\newblock \emph{ACM Transactions on Graphics, (Proc. SIGGRAPH Asia)},
  34\penalty0 (6):\penalty0 248:1--248:16, 2015.

\bibitem[Ma et~al.(2008)Ma, Jones, Chiang, Hawkins, Frederiksen, Peers,
  Vukovic, Ouhyoung, and Debevec]{ma2008facial}
Wan-Chun Ma, Andrew Jones, Jen-Yuan Chiang, Tim Hawkins, Sune Frederiksen,
  Pieter Peers, Marko Vukovic, Ming Ouhyoung, and Paul Debevec.
\newblock Facial performance synthesis using deformation-driven polynomial
  displacement maps.
\newblock \emph{{Transactions on Graphics (TOG)}}, 27\penalty0 (5):\penalty0
  1--10, 2008.

\bibitem[Metzer et~al.(2023)Metzer, Richardson, Patashnik, Giryes, and
  Cohen-Or]{metzer2022latentnerf}
Gal Metzer, Elad Richardson, Or Patashnik, Raja Giryes, and Daniel Cohen-Or.
\newblock Latent-nerf for shape-guided generation of 3d shapes and textures.
\newblock In \emph{{Computer Vision and Pattern Recognition (CVPR)}}, 2023.

\bibitem[Michel et~al.(2022)Michel, Bar-On, Liu, Benaim, and
  Hanocka]{text2mesh}
Oscar Michel, Roi Bar-On, Richard Liu, Sagie Benaim, and Rana Hanocka.
\newblock {Text2Mesh: Text-Driven Neural Stylization for Meshes}.
\newblock In \emph{{Computer Vision and Pattern Recognition (CVPR)}}, 2022.

\bibitem[Mildenhall et~al.(2020)Mildenhall, Srinivasan, Tancik, Barron,
  Ramamoorthi, and Ng]{mildenhall2020nerf}
Ben Mildenhall, Pratul~P. Srinivasan, Matthew Tancik, Jonathan~T. Barron, Ravi
  Ramamoorthi, and Ren Ng.
\newblock Nerf: Representing scenes as neural radiance fields for view
  synthesis.
\newblock In \emph{{European Conference on Computer Vision (ECCV)}}, 2020.

\bibitem[Mohammad~Khalid et~al.(2022)Mohammad~Khalid, Xie, Belilovsky, and
  Popa]{mohammad2022clipmesh}
Nasir Mohammad~Khalid, Tianhao Xie, Eugene Belilovsky, and Tiberiu Popa.
\newblock Clip-mesh: Generating textured meshes from text using pretrained
  image-text models.
\newblock In \emph{SIGGRAPH Asia 2022 Conference Papers}, pages 1--8, 2022.

\bibitem[Mou et~al.(2023)Mou, Wang, Xie, Zhang, Qi, Shan, and Qie]{mou2023t2i}
Chong Mou, Xintao Wang, Liangbin Xie, Jian Zhang, Zhongang Qi, Ying Shan, and
  Xiaohu Qie.
\newblock T2i-adapter: Learning adapters to dig out more controllable ability
  for text-to-image diffusion models.
\newblock \emph{arXiv preprint:2302.08453}, 2023.

\bibitem[Park et~al.(2019)Park, Florence, Straub, Newcombe, and
  Lovegrove]{park2019deepsdf}
Jeong~Joon Park, Peter Florence, Julian Straub, Richard Newcombe, and Steven
  Lovegrove.
\newblock Deepsdf: Learning continuous signed distance functions for shape
  representation.
\newblock In \emph{{Computer Vision and Pattern Recognition (CVPR)}}, pages
  165--174, 2019.

\bibitem[Pavlakos et~al.(2019)Pavlakos, Choutas, Ghorbani, Bolkart, Osman,
  Tzionas, and Black]{SMPL-X:2019}
Georgios Pavlakos, Vasileios Choutas, Nima Ghorbani, Timo Bolkart, Ahmed A.~A.
  Osman, Dimitrios Tzionas, and Michael~J. Black.
\newblock {Expressive Body Capture: 3D Hands, Face, and Body from a Single
  Image}.
\newblock In \emph{{Computer Vision and Pattern Recognition (CVPR)}}, 2019.

\bibitem[Poole et~al.(2023)Poole, Jain, Barron, and
  Mildenhall]{poole2022dreamfusion}
Ben Poole, Ajay Jain, Jonathan~T Barron, and Ben Mildenhall.
\newblock {DreamFusion}: Text-to-3d using 2d diffusion.
\newblock In \emph{{International Conference on Learning Representations
  (ICLR)}}, 2023.

\bibitem[Qiu et~al.(2023)Qiu, Liu, Feng, Xue, Feng, Liu, Zhang, Weller, and
  Schölkopf]{oft2023}
Zeju Qiu, Weiyang Liu, Haiwen Feng, Yuxuan Xue, Yao Feng, Zhen Liu, Dan Zhang,
  Adrian Weller, and Bernhard Schölkopf.
\newblock {Controlling Text-to-Image Diffusion by Orthogonal Finetuning}.
\newblock \emph{arXiv preprint:2306.07280}, 2023.

\bibitem[Raffel et~al.(2020)Raffel, Shazeer, Roberts, Lee, Narang, Matena,
  Zhou, Li, and Liu]{raffel2020exploring}
Colin Raffel, Noam Shazeer, Adam Roberts, Katherine Lee, Sharan Narang, Michael
  Matena, Yanqi Zhou, Wei Li, and Peter~J Liu.
\newblock Exploring the limits of transfer learning with a unified text-to-text
  transformer.
\newblock \emph{The Journal of Machine Learning Research}, 21\penalty0
  (1):\penalty0 5485--5551, 2020.

\bibitem[Ramesh et~al.(2021)Ramesh, Pavlov, Goh, Gray, Voss, Radford, Chen, and
  Sutskever]{ramesh2021zero}
Aditya Ramesh, Mikhail Pavlov, Gabriel Goh, Scott Gray, Chelsea Voss, Alec
  Radford, Mark Chen, and Ilya Sutskever.
\newblock Zero-shot text-to-image generation.
\newblock In \emph{{International Conference on Machine Learning (ICML)}},
  pages 8821--8831. PMLR, 2021.

\bibitem[Richardson et~al.(2023)Richardson, Metzer, Alaluf, Giryes, and
  Cohen-Or]{richardson2023texture}
Elad Richardson, Gal Metzer, Yuval Alaluf, Raja Giryes, and Daniel Cohen-Or.
\newblock Texture: Text-guided texturing of 3d shapes.
\newblock \emph{arXiv preprint:2302.01721}, 2023.

\bibitem[Rombach et~al.(2022)Rombach, Blattmann, Lorenz, Esser, and
  Ommer]{rombach2022high}
Robin Rombach, Andreas Blattmann, Dominik Lorenz, Patrick Esser, and Bj{\"o}rn
  Ommer.
\newblock High-resolution image synthesis with latent diffusion models.
\newblock In \emph{{Computer Vision and Pattern Recognition (CVPR)}}, pages
  10684--10695, 2022.

\bibitem[Ruiz et~al.(2023)Ruiz, Li, Jampani, Pritch, Rubinstein, and
  Aberman]{ruiz2022dreambooth}
Nataniel Ruiz, Yuanzhen Li, Varun Jampani, Yael Pritch, Michael Rubinstein, and
  Kfir Aberman.
\newblock Dreambooth: Fine tuning text-to-image diffusion models for
  subject-driven generation.
\newblock In \emph{{Computer Vision and Pattern Recognition (CVPR)}}, 2023.

\bibitem[Saharia et~al.(2022)Saharia, Chan, Saxena, Li, Whang, Denton,
  Ghasemipour, Gontijo~Lopes, Karagol~Ayan, Salimans,
  et~al.]{saharia2022photorealistic}
Chitwan Saharia, William Chan, Saurabh Saxena, Lala Li, Jay Whang, Emily~L
  Denton, Kamyar Ghasemipour, Raphael Gontijo~Lopes, Burcu Karagol~Ayan, Tim
  Salimans, et~al.
\newblock Photorealistic text-to-image diffusion models with deep language
  understanding.
\newblock In \emph{{Conference on Neural Information Processing Systems
  (NeurIPS)}}, pages 36479--36494, 2022.

\bibitem[Saito et~al.(2019)Saito, Huang, Natsume, Morishima, Kanazawa, and
  Li]{saito2019pifu}
Shunsuke Saito, Zeng Huang, Ryota Natsume, Shigeo Morishima, Angjoo Kanazawa,
  and Hao Li.
\newblock Pifu: Pixel-aligned implicit function for high-resolution clothed
  human digitization.
\newblock In \emph{{International Conference on Computer Vision ({ICCV})}},
  2019.

\bibitem[Saito et~al.(2020)Saito, Simon, Saragih, and Joo]{saito2020pifuhd}
Shunsuke Saito, Tomas Simon, Jason Saragih, and Hanbyul Joo.
\newblock Pifuhd: Multi-level pixel-aligned implicit function for
  high-resolution 3d human digitization.
\newblock In \emph{CVPR}, 2020.

\bibitem[Sanghi et~al.(2022)Sanghi, Chu, Lambourne, Wang, Cheng, Fumero, and
  Malekshan]{sanghi2022clipforge}
Aditya Sanghi, Hang Chu, Joseph~G Lambourne, Ye Wang, Chin-Yi Cheng, Marco
  Fumero, and Kamal~Rahimi Malekshan.
\newblock Clip-forge: Towards zero-shot text-to-shape generation.
\newblock In \emph{{Computer Vision and Pattern Recognition (CVPR)}}, pages
  18603--18613, 2022.

\bibitem[Santesteban et~al.(2021)Santesteban, Thuerey, Otaduy, and
  Casas]{santesteban2021garmentcollisions}
Igor Santesteban, Nils Thuerey, Miguel~A Otaduy, and Dan Casas.
\newblock {Self-Supervised Collision Handling via Generative 3D Garment Models
  for Virtual Try-On}.
\newblock In \emph{{Computer Vision and Pattern Recognition (CVPR)}}, 2021.

\bibitem[Santesteban et~al.(2022)Santesteban, Otaduy, Thuerey, and
  Casas]{santesteban2021ulnefs}
Igor Santesteban, Miguel~A. Otaduy, Nils Thuerey, and Dan Casas.
\newblock {ULNeF}: Untangled layered neural fields for mix-and-match virtual
  try-on.
\newblock In \emph{{Conference on Neural Information Processing Systems
  (NeurIPS)}}, 2022.

\bibitem[Shafir et~al.(2023)Shafir, Tevet, Kapon, and Bermano]{shafir2023human}
Yonatan Shafir, Guy Tevet, Roy Kapon, and Amit~H Bermano.
\newblock Human motion diffusion as a generative prior.
\newblock \emph{arXiv preprint arXiv:2303.01418}, 2023.

\bibitem[Shao et~al.(2022)Shao, Zheng, Zhang, Sun, and
  Liu]{shao2022diffusionstereo}
Ruizhi Shao, Zerong Zheng, Hongwen Zhang, Jingxiang Sun, and Yebin Liu.
\newblock Diffustereo: High quality human reconstruction via diffusion-based
  stereo using sparse cameras.
\newblock In \emph{{European Conference on Computer Vision (ECCV)}}, 2022.

\bibitem[Shen et~al.(2021)Shen, Gao, Yin, Liu, and Fidler]{shen2021deep}
Tianchang Shen, Jun Gao, Kangxue Yin, Ming-Yu Liu, and Sanja Fidler.
\newblock Deep marching tetrahedra: a hybrid representation for high-resolution
  3d shape synthesis.
\newblock In \emph{{Conference on Neural Information Processing Systems
  (NeurIPS)}}, pages 6087--6101, 2021.

\bibitem[Sklyarova et~al.(2023)Sklyarova, Chelishev, Dogaru, Medvedev,
  Lempitsky, and Zakharov]{sklyarova2023neural}
Vanessa Sklyarova, Jenya Chelishev, Andreea Dogaru, Igor Medvedev, Victor
  Lempitsky, and Egor Zakharov.
\newblock {Neural Haircut: Prior-Guided Strand-Based Hair Reconstruction}.
\newblock In \emph{{International Conference on Computer Vision ({ICCV})}},
  2023.

\bibitem[Sohl-Dickstein et~al.(2015)Sohl-Dickstein, Weiss, Maheswaranathan, and
  Ganguli]{sohl2015deep}
Jascha Sohl-Dickstein, Eric Weiss, Niru Maheswaranathan, and Surya Ganguli.
\newblock Deep unsupervised learning using nonequilibrium thermodynamics.
\newblock In \emph{International conference on machine learning}, pages
  2256--2265. PMLR, 2015.

\bibitem[Song et~al.(2021)Song, Meng, and Ermon]{song2020denoising}
Jiaming Song, Chenlin Meng, and Stefano Ermon.
\newblock Denoising diffusion implicit models.
\newblock In \emph{{International Conference on Learning Representations
  (ICLR)}}, 2021.

\bibitem[Song and Ermon(2019)]{song2019generative}
Yang Song and Stefano Ermon.
\newblock Generative modeling by estimating gradients of the data distribution.
\newblock In \emph{{Conference on Neural Information Processing Systems
  (NeurIPS)}}, 2019.

\bibitem[Tevet et~al.(2023)Tevet, Raab, Gordon, Shafir, Cohen-or, and
  Bermano]{tevet2023human}
Guy Tevet, Sigal Raab, Brian Gordon, Yoni Shafir, Daniel Cohen-or, and
  Amit~Haim Bermano.
\newblock Human motion diffusion model.
\newblock In \emph{{International Conference on Learning Representations
  (ICLR)}}, 2023.

\bibitem[tts()]{tts}
tts.
\newblock \url{https://play.ht}, 2023.

\bibitem[Wang et~al.(2022)Wang, Chai, He, Chen, and Liao]{wang2022clipnerf}
Can Wang, Menglei Chai, Mingming He, Dongdong Chen, and Jing Liao.
\newblock Clip-nerf: Text-and-image driven manipulation of neural radiance
  fields.
\newblock In \emph{{Computer Vision and Pattern Recognition (CVPR)}}, pages
  3835--3844, 2022.

\bibitem[Wang et~al.(2021)Wang, Liu, Liu, Theobalt, Komura, and
  Wang]{wangNeuSLearningNeural2021}
Peng Wang, Lingjie Liu, Yuan Liu, Christian Theobalt, Taku Komura, and Wenping
  Wang.
\newblock {NeuS: Learning Neural Implicit Surfaces by Volume Rendering for
  Multi-view Reconstruction}.
\newblock In \emph{{Conference on Neural Information Processing Systems
  (NeurIPS)}}, 2021.

\bibitem[Wang et~al.(2023{\natexlab{a}})Wang, Zhang, Zhang, Gu, Bao,
  Baltrusaitis, Shen, Chen, Wen, Chen, et~al.]{wang2023rodin}
Tengfei Wang, Bo Zhang, Ting Zhang, Shuyang Gu, Jianmin Bao, Tadas
  Baltrusaitis, Jingjing Shen, Dong Chen, Fang Wen, Qifeng Chen, et~al.
\newblock Rodin: A generative model for sculpting 3d digital avatars using
  diffusion.
\newblock In \emph{{Computer Vision and Pattern Recognition (CVPR)}}, pages
  4563--4573, 2023{\natexlab{a}}.

\bibitem[Wang et~al.(2023{\natexlab{b}})Wang, Lu, Wang, Bao, Li, Su, and
  Zhu]{wang2023prolificdreamer}
Zhengyi Wang, Cheng Lu, Yikai Wang, Fan Bao, Chongxuan Li, Hang Su, and Jun
  Zhu.
\newblock Prolificdreamer: High-fidelity and diverse text-to-3d generation with
  variational score distillation.
\newblock \emph{arXiv preprint arXiv:2305.16213}, 2023{\natexlab{b}}.

\bibitem[Welling and Teh(2011)]{welling2011bayesian}
Max Welling and Yee~W Teh.
\newblock Bayesian learning via stochastic gradient langevin dynamics.
\newblock In \emph{{International Conference on Machine Learning (ICML)}},
  pages 681--688, 2011.

\bibitem[Xiu et~al.(2022)Xiu, Yang, Tzionas, and Black]{xiu2022icon}
Yuliang Xiu, Jinlong Yang, Dimitrios Tzionas, and Michael~J. Black.
\newblock {ICON}: {I}mplicit {C}lothed humans {O}btained from {N}ormals.
\newblock In \emph{{Computer Vision and Pattern Recognition (CVPR)}}, 2022.

\bibitem[Xiu et~al.(2023)Xiu, Yang, Cao, Tzionas, and Black]{xiu2023econ}
Yuliang Xiu, Jinlong Yang, Xu Cao, Dimitrios Tzionas, and Michael~J. Black.
\newblock {ECON: Explicit Clothed humans Optimized via Normal integration}.
\newblock In \emph{{Computer Vision and Pattern Recognition (CVPR)}}, 2023.

\bibitem[Xu et~al.(2023)Xu, Wang, Cheng, Cao, Shan, Qie, and
  Gao]{xu2022dream3d}
Jiale Xu, Xintao Wang, Weihao Cheng, Yan-Pei Cao, Ying Shan, Xiaohu Qie, and
  Shenghua Gao.
\newblock Dream3d: Zero-shot text-to-3d synthesis using 3d shape prior and
  text-to-image diffusion models.
\newblock In \emph{{Computer Vision and Pattern Recognition (CVPR)}}, 2023.

\bibitem[Yang et~al.(2023)Yang, Luo, Xiu, Wang, Xu, and Fan]{yang2023dif}
Xueting Yang, Yihao Luo, Yuliang Xiu, Wei Wang, Hao Xu, and Zhaoxin Fan.
\newblock {D-IF: Uncertainty-aware Human Digitization via Implicit Distribution
  Field}.
\newblock In \emph{{International Conference on Computer Vision ({ICCV})}},
  2023.

\bibitem[Yi et~al.(2023{\natexlab{a}})Yi, Huang, Tripathi, Hering, Thies, and
  Black]{yi2023mime}
Hongwei Yi, Chun-Hao~P. Huang, Shashank Tripathi, Lea Hering, Justus Thies, and
  Michael~J. Black.
\newblock {MIME}: Human-aware {3D} scene generation.
\newblock In \emph{{Computer Vision and Pattern Recognition (CVPR)}},
  2023{\natexlab{a}}.

\bibitem[Yi et~al.(2023{\natexlab{b}})Yi, Liang, Liu, Cao, Wen, Bolkart, Tao,
  and Black]{yi2022talkshow}
Hongwei Yi, Hualin Liang, Yifei Liu, Qiong Cao, Yandong Wen, Timo Bolkart,
  Dacheng Tao, and Michael~J Black.
\newblock Generating holistic 3d human motion from speech.
\newblock In \emph{{Computer Vision and Pattern Recognition (CVPR)}},
  2023{\natexlab{b}}.

\bibitem[Zhang et~al.(2023{\natexlab{a}})Zhang, Chen, Yang, Qu, Wang, Chen,
  Long, Zhu, Du, and Zheng]{zhang2023avatarverse}
Huichao Zhang, Bowen Chen, Hao Yang, Liao Qu, Xu Wang, Li Chen, Chao Long,
  Feida Zhu, Kang Du, and Min Zheng.
\newblock Avatarverse: High-quality \& stable 3d avatar creation from text and
  pose.
\newblock \emph{arXiv preprint:2308.03610}, 2023{\natexlab{a}}.

\bibitem[Zhang and Agrawala(2023)]{zhang2023adding}
Lvmin Zhang and Maneesh Agrawala.
\newblock Adding conditional control to text-to-image diffusion models, 2023.

\bibitem[Zhang et~al.(2023{\natexlab{b}})Zhang, Qiu, Lin, Zhang, Shi, Yang,
  Shi, Yang, Xu, and Yu]{zhang2023dreamface}
Longwen Zhang, Qiwei Qiu, Hongyang Lin, Qixuan Zhang, Cheng Shi, Wei Yang, Ye
  Shi, Sibei Yang, Lan Xu, and Jingyi Yu.
\newblock Dreamface: Progressive generation of animatable 3d faces under text
  guidance.
\newblock \emph{arXiv preprint arXiv:2304.03117}, 2023{\natexlab{b}}.

\bibitem[Zhao et~al.(2023)Zhao, Li, Hu, Li, Zou, Shi, and Fan]{zhao2023zero}
Rui Zhao, Wei Li, Zhipeng Hu, Lincheng Li, Zhengxia Zou, Zhenwei Shi, and
  Changjie Fan.
\newblock Zero-shot text-to-parameter translation for game character
  auto-creation.
\newblock In \emph{{Computer Vision and Pattern Recognition (CVPR)}}, pages
  21013--21023, 2023.

\bibitem[Zheng et~al.(2021)Zheng, Shao, Zhang, Yu, Zheng, Dai, and
  Liu]{zheng2021deepmulticap}
Yang Zheng, Ruizhi Shao, Yuxiang Zhang, Tao Yu, Zerong Zheng, Qionghai Dai, and
  Yebin Liu.
\newblock Deepmulticap: Performance capture of multiple characters using sparse
  multiview cameras.
\newblock In \emph{{International Conference on Computer Vision ({ICCV})}},
  2021.

\bibitem[Zheng et~al.(2023)Zheng, Zhao, Zhang, Liu, and
  Liu]{zheng2023avatarrex}
Zerong Zheng, Xiaochen Zhao, Hongwen Zhang, Boning Liu, and Yebin Liu.
\newblock Avatarrex: Real-time expressive full-body avatars.
\newblock \emph{ACM Transactions on Graphics (TOG)}, 42\penalty0 (4), 2023.

\bibitem[Zhu et~al.(2022)Zhu, Qiu, Qiu, and Han]{zhu2022registering}
Heming Zhu, Lingteng Qiu, Yuda Qiu, and Xiaoguang Han.
\newblock {Registering explicit to implicit: Towards high-fidelity garment mesh
  reconstruction from single images}.
\newblock In \emph{{Computer Vision and Pattern Recognition (CVPR)}}, pages
  3845--3854, 2022.

\bibitem[Zhu et~al.(2020)Zhu, Rematas, Curless, Seitz, and
  Kemelmacher-Shlizerman]{zhu_2020_eccv_nba}
Luyang Zhu, Konstantinos Rematas, Brian Curless, Steve Seitz, and Ira
  Kemelmacher-Shlizerman.
\newblock Reconstructing nba players.
\newblock In \emph{{European Conference on Computer Vision (ECCV)}}, 2020.

\end{thebibliography}
}

\clearpage

\begin{appendices} 
\label{appendices}

\section{Additional Qualitative Comparisons}
We provide additional qualitative comparisons with existing baselines in three categories:
head avatar generation, 
full-body iconic and customized avatar generation.

\begin{figure}[t!]
\centering 
\includegraphics[ trim=000mm 008mm 000mm 000mm, clip=True, 
width=\linewidth]{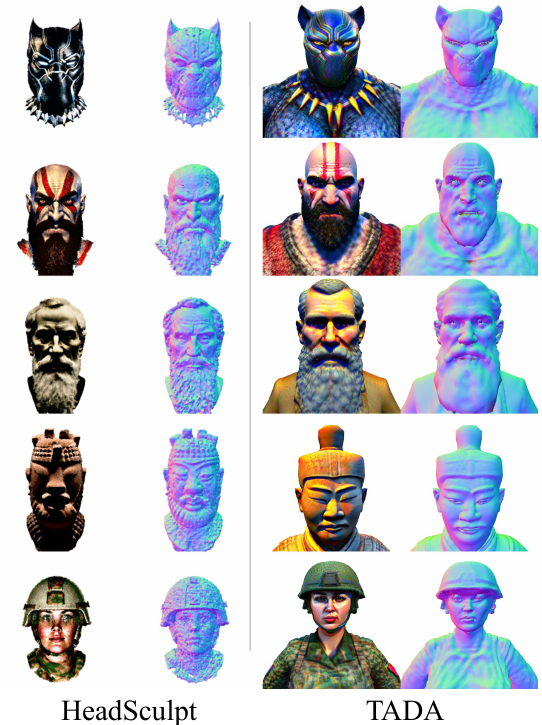}
\begin{tabularx}{\linewidth}{
 >{\centering\arraybackslash}X
 >{\centering\arraybackslash}X
}
HeadSculpt~\cite{han2023headsculpt} &
TADA (Ours)
\end{tabularx}
\caption{Qualitative comparison of our method TADA with HeadSculpt~\cite{han2023headsculpt} on the task of head avatar generation.}
\label{fig:compare_sup_hs}
\end{figure}

\smallskip
\noindent \textbf{Head Avatar Generation.}
In \fref{fig:compare_sup_hs}, we provide examples generated by HeadSculpt~\cite{han2023headsculpt} and our method TADA.
Unlike HeadSculpt often generates noisy geometry and suffers from inconsistency between texture and geometry, TADA could produce high-quality character geometries with well-aligned textures.

\begin{figure*}[t!]
    \centering 
    \includegraphics[trim=000mm 007mm 000mm 000mm, clip=True,width=\linewidth]{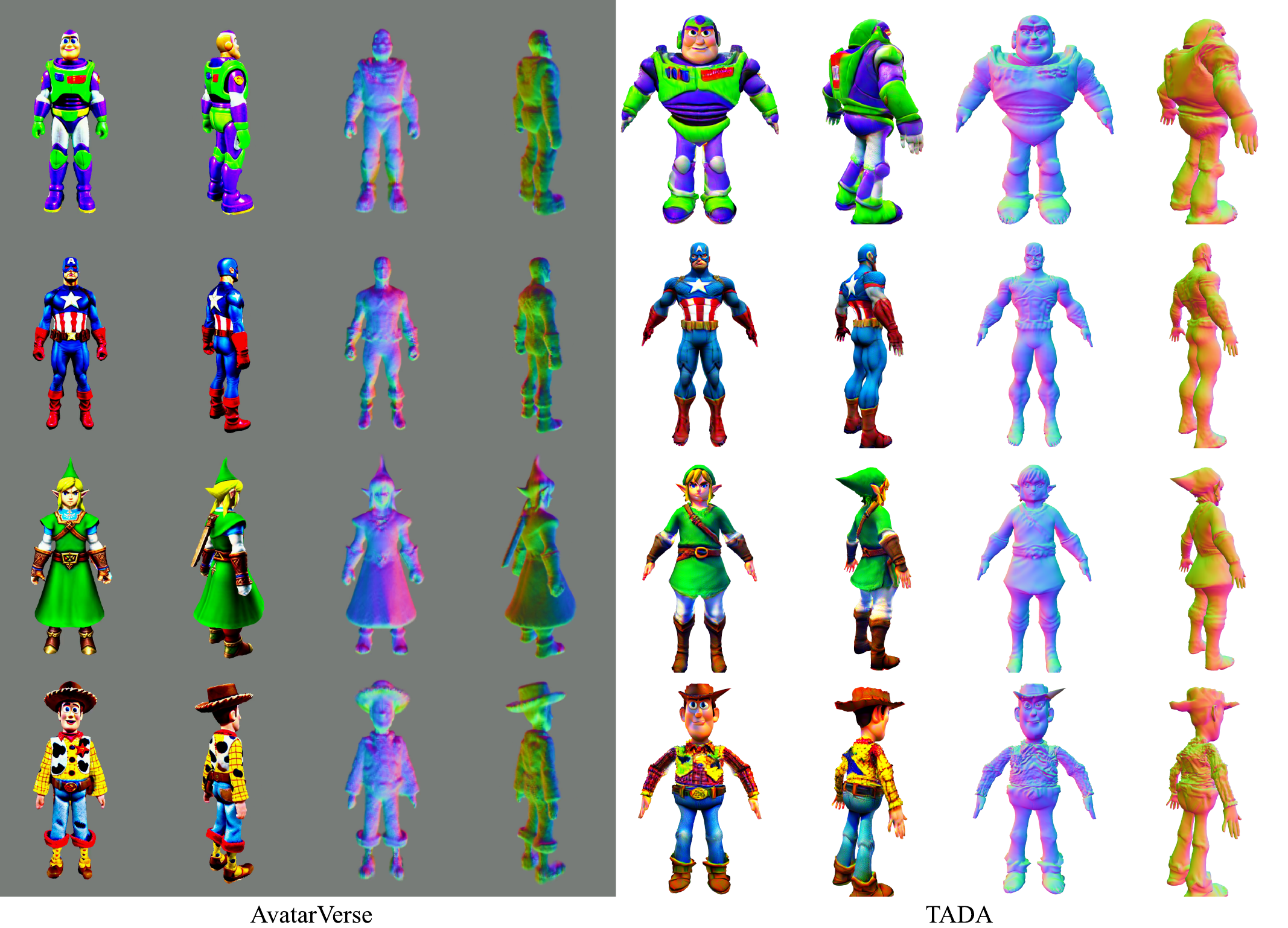}
    \begin{tabularx}{\linewidth}{
 >{\centering\arraybackslash}X
 >{\centering\arraybackslash}X
}
AvatarVerse \cite{zhang2023avatarverse} &
TADA (Ours)
\end{tabularx}
    \caption{
    \textbf{Qualitative comparison of the full-body iconic avatar generation with AvatarVerse \cite{zhang2023avatarverse}.} Left is from AvatarVerse. Right is ours. 
    The generated avatars from AvatarVerse lack high-quality details in geometry, especially on the face region. In contrast, TADA generates high-quality meshes with well-aligned textures.
    }
    \label{fig:compare_sup_dv}
\end{figure*}

\begin{figure*}[t]
\centering 
\includegraphics[trim=000mm 008mm 000mm 000mm, clip=True,  
width=\linewidth]{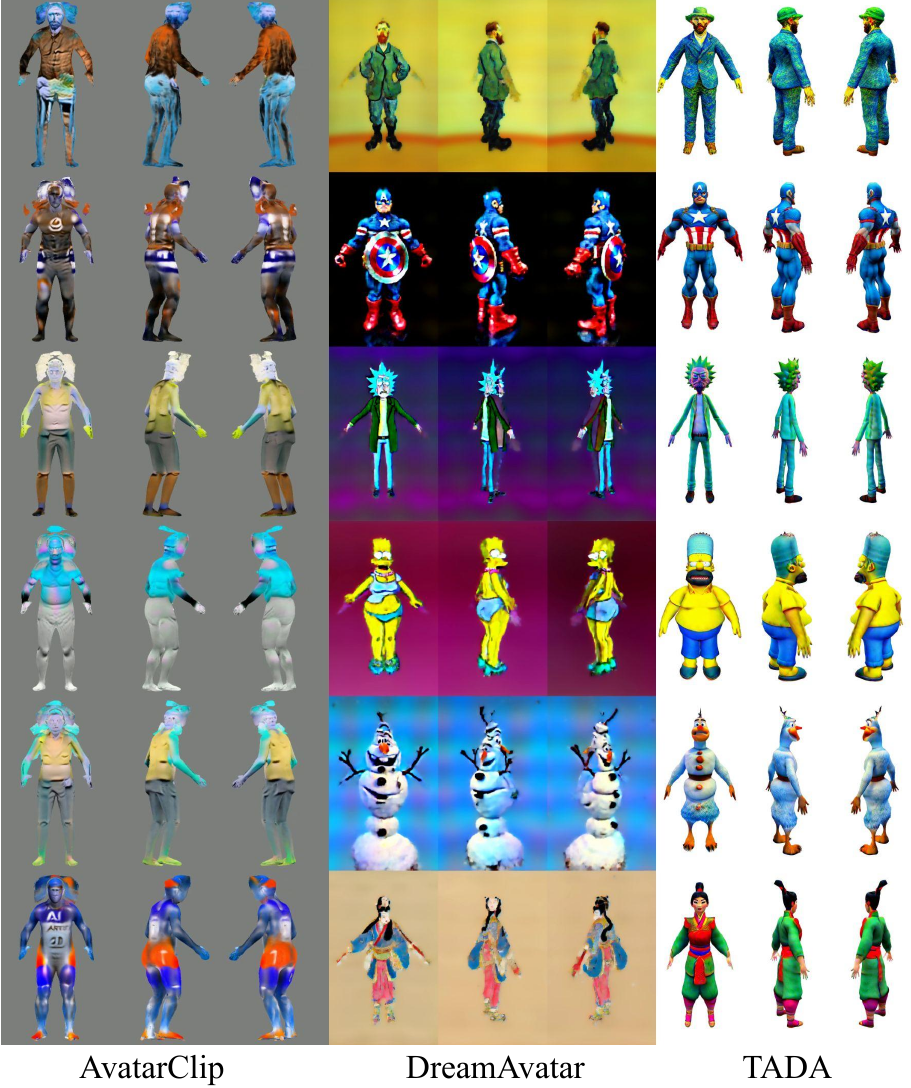}
\begin{tabularx}{\linewidth}{
 >{\centering\arraybackslash}X
 >{\centering\arraybackslash}X
 >{\centering\arraybackslash}X
}
AvatarCLIP~\cite{hong2022avatarclip} &
DreamAvatar~\cite{cao2023dreamavatar} &
TADA (Ours)
\end{tabularx}
\caption{Qualitative comparison against DreamAvatar~\cite{cao2023dreamavatar} and AvatarClip~\cite{hong2022avatarclip} on the full-body iconic avatar generation.
}
\label{fig:compare_sup_ac}
\end{figure*}

\smallskip
\noindent \textbf{Full-body Iconic Avatar Generation.}
We conduct a comprehensive comparison of the full-body iconic avatar generation with existing methods \cite{poole2022dreamfusion, cao2023dreamavatar, huang2023dreamwaltz, zhang2023avatarverse, hong2022avatarclip} in \cref{fig:teaser,fig:compare_sup_dv,fig:compare_sup_ac}.
NeRF-based methods~\cite{poole2022dreamfusion, cao2023dreamavatar, zhang2023avatarverse, huang2023dreamwaltz} tends to generate low-quality geometric fields, which are not compatible with existing traditional CG workflows, such as rasterization and animation.
The mesh-based method AvatarClip~\cite{hong2022avatarclip} tends to generate minimal clothed geometry, and low-quality texture with severe artifacts (see \fref{fig:compare_sup_ac}).
In contrast, TADA has the capability to produce a wide range of characters with superior geometries and well-aligned textures. These outputs seamlessly integrate with conventional graphics workflows, making them readily suitable for animation and rendering.

\begin{figure*}[t]
\centering 
\includegraphics[ trim=000mm 006mm 000mm 000mm, clip=True,
width=\linewidth]{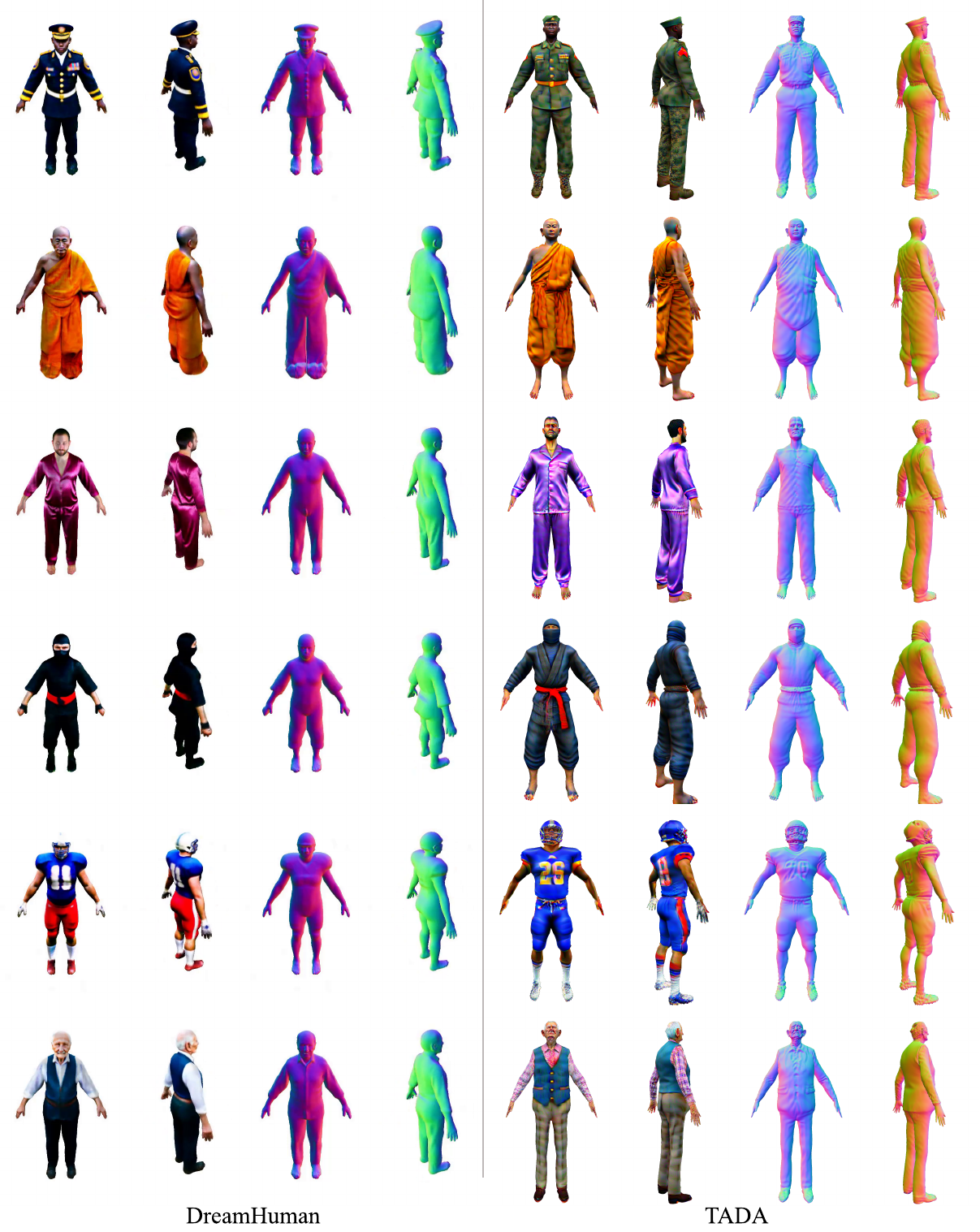}
\begin{tabularx}{\linewidth}{
 >{\centering\arraybackslash}X
 >{\centering\arraybackslash}X
}
DreamHuman \cite{kolotouros2023dreamhuman} &
TADA (Ours)
\end{tabularx}
\caption{Qualitative comparison with DreamHuman \cite{kolotouros2023dreamhuman} on the full-body customized avatar generation.}
\label{fig:compare_sup_dh}
\end{figure*}

\smallskip
\noindent \textbf{Full-body Customized Avatars Generation.}
To explore the generalization of our method, we compare TADA with DreamHuman~\cite{kolotouros2023dreamhuman} on the task of full-body customized avatars generation.
%
In \fref{fig:compare_sup_dh}, we can see that TADA generates high-quality characters with more consistent geometry and texture than DreamHuman.
%

\section{Implementation details}
\label{sec:Implementation_details}

We select camera positions ($r, \theta, \phi$) in a spherical coordinate system, where $r, \theta, \text{and } \phi$ denote the radius, elevation, and azimuth angle, respectively.
During the optimization, the virtual cameras are positioned as: 1) \textbf{full-body mode}: camera with full-body field of view (FOV), and 2) \textbf{head-mode}: zoom-in camera focusing the head.
The head-mode camera is sampled with a probability of 30\%, and full-body model with 70\%.
We sample $\theta_b$ values within the range of $[60^\circ, 90^\circ]$ under full-body mode.
Conversely, for the head view $\theta_h$, we opt for values from the range of $[75^\circ, 85^\circ]$. 
Additionally, we sample $\phi$ within the intervals of $[-180^\circ, 180^\circ]$ for the full body, and $[-30^\circ, 30^\circ]$ for the head. 
In each iteration, the camera radius is adjusted according to the body height and the head size. 

\begin{figure*}[t!]
    \centering
    \includegraphics[trim=000mm 015mm 000mm 000mm, clip=True,width=\linewidth]{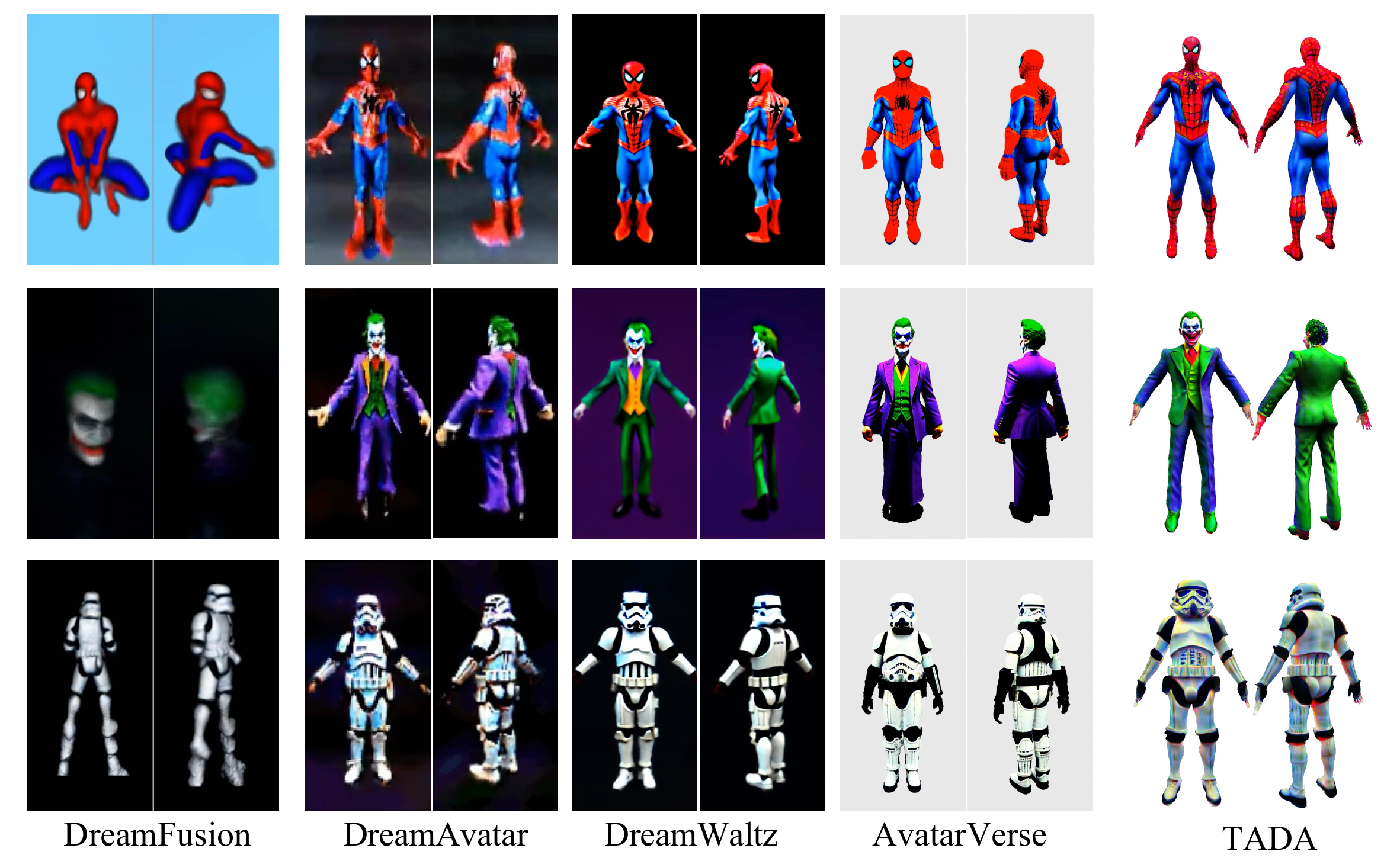}
    \begin{tabularx}{\linewidth}{
 >{\centering\arraybackslash}X
 >{\centering\arraybackslash}X
 >{\centering\arraybackslash}X
 >{\centering\arraybackslash}X
 >{\centering\arraybackslash}X
}
DreamFusion~\cite{poole2022dreamfusion} &
DreamAvatar~\cite{cao2023dreamavatar} &
DreamWaltz~\cite{huang2023dreamwaltz} &
AvatarVerse \cite{zhang2023avatarverse} &
TADA (Ours)
\end{tabularx}
    \captionof{figure}{
    \textbf{Qualitative comparison on full-body avatar generation of icons.} The prompts (top $\rightarrow$ down) are \textit{``Spiderman'', ``Joker'', ``Stormtrooper''}. Compared with the baselines \cite{poole2022dreamfusion, cao2023dreamavatar, huang2023dreamwaltz, zhang2023avatarverse}, \modelname generates a higher quality in terms of both geometry and texture. 
    }
    \label{fig:teaser}
\end{figure*}

\clearpage
\section{Prompts used by ChatGPT}
\label{sec: chatgpt}
%
%
We generate characters of three groups: celebrity, characters in movies \& anime (fictional characters), and general job descriptions.
We also provide the prompts used to imagine outfits for Virtual Try-on application.
In the following, we list the prompts used for generation.

\medskip
\noindent \textbf{Celebrities.}
We ask ChatGPT to output the names of superstars, scientists, businessmen and presidents.
Here are the corresponding generated prompts:
%
{\tiny
\begin{verbatim}
Donald Trump
Abraham Lincoln
Barack Obama
Hilary Clinton
Yao Ming
Kobe Bryant
Messi
Bruce Lee
Steven Paul Jobs
Mark Elliot Zuckerberg
Joe Biden
Bill Gates
Warren Buffett
Elon Musk
Jeff Bezos
Jack Ma
Albert Einstein
Marie Curie
Stephen Hawking
Vincent van Gogh
Michelangelo
Wolfgang Amadeus Mozart
Ludwig van Beethoven
Michael Jackson
Kim Kardashian
\end{verbatim} 
}

\noindent \textbf{Fictional Characters.}
We collect superheroes, Disney princesses, and characters in films such as Forzen, Aladdin, South Park, Simpson, Rick and Morty, \etc.
%
{\tiny
\begin{verbatim}
superman
Deadpool
Batman
Ant-man
spiderman
Iron Man
Captain America
Woody in Toy Story
Buzz Lightyear in Toy Story
Elsa in Frozen
Anna in Frozen
Kristoff in Frozen
Aladdin in Aladdin
Jasmine in Aladdin
Mulan in Mulan
Olaf in Frozen
Jessie in Toy Story
Sun Wukong
groot
Moana in Moana
Judy Hopps in Zootopia
Goku in Dragon Ball series
Naruto Uzumaki in Naruto series
Luffy in One Piece
Kiki in Kiki's Delivery Service
San in Princess Mononoke
Eric Cartman in South Park: Bigger, Longer & Uncut
Dipper Pines in Gravity Falls
Mabel Pines in Gravity Falls
Rick Sanchez in Rick and Morty
Morty Smith in Rick and Morty
Stan Marsh in South Park: Bigger, Longer & Uncut
Grunkle Stan in Gravity Falls
Soos Ramirez in Gravity Falls
Kyle Broflovski in South Park: Bigger, Longer & Uncut
Kenny McCormick in South Park: Bigger, Longer & Uncut
Bojack Horseman in Bojack Horseman
Diane Nguyen in Bojack Horseman
Ariel in The Little Mermaid
Belle in Beauty and the Beast
Homer Simpson in The Simpsons
Marge Simpson in The Simpsons
Bart Simpson in The Simpsons
Lisa Simpson in The Simpsons
astronaut
Michelangelo style statue of an astronaut
A highly detailed stone bust of Theodoros Kolokotronis
Captain Jack Sparrow from Pirates of the Caribbean
Mobile suit Gundam
Harry Potter, adult
Optimus Prime, transformer
Ronald Weasley, adult
stormtrooper
Casino Dealer
Ancient Prince of India
Nick Fury
Captain Marvel
robot
Thor
Doctor Strange
Snow White
Cinderella
Crayon Shin-chan
Misae Nohara
\end{verbatim} 
}

\noindent \textbf{General Job Descriptions.}
In addition, we ask GPT to generate general job descriptions such as policeman, doctor, chef and \etc.
{\tiny
\begin{verbatim}
gardener
warrior
knight
witch
wizard
Joker
Clown
Chef
Nurse
Dentist
Doctor
Artist
Carpenter
Farmer
Athlete
Pilot
Firefighter
Police officer
astronaut
\end{verbatim} 
}

\noindent \textbf{Imagined Outfits for Virtual Try-on.}
Here is an example, how we use ChatGPT to generate different outfit descriptions for Barack Obama. 
{\tiny
\begin{verbatim}
We: please generate 50 prompts describing Obama wearing different clothes

----------------------------------------------------------------

ChatGPT: 
Sure, here are 50 writing prompts that describe Barack Obama
wearing different clothes:

Barack Obama wearing a sharp navy blue suit and a red tie.
Barack Obama wearing a casual grey t-shirt and jeans.
Barack Obama wearing a black leather jacket and aviator sunglasses.
Barack Obama wearing a traditional African dashiki.
Barack Obama wearing a classic American baseball cap and polo shirt.
Barack Obama wearing a white dress shirt and black dress pants.
Barack Obama wearing a bright yellow raincoat and rain boots.
Barack Obama wearing a navy blue hoodie and sneakers.
Barack Obama wearing a tan safari jacket and khaki pants.
Barack Obama wearing a red and black plaid flannel shirt and jeans.
Barack Obama wearing a crisp white dress shirt and a black suit.
Barack Obama wearing a light blue chambray shirt and khaki shorts.
Barack Obama wearing a navy blue blazer and chinos.
Barack Obama wearing a black and white houndstooth suit and tie.
Barack Obama wearing a green military-style jacket and combat boots.
Barack Obama wearing a purple dress shirt and grey suit.
Barack Obama wearing a denim shirt and khaki pants.
Barack Obama wearing a pink dress shirt and navy blue suit.
Barack Obama wearing a brown leather jacket and brown boots.
Barack Obama wearing a yellow polo shirt and white shorts.
Barack Obama wearing a grey sweatshirt and athletic pants.
Barack Obama wearing a black leather bomber jacket and jeans.
Barack Obama wearing a white t-shirt and denim jacket.
Barack Obama wearing a burgundy sweater and dress pants.
Barack Obama wearing a black and white checkered suit and tie.
Barack Obama wearing a green army jacket and cargo pants.
Barack Obama wearing a navy blue blazer and grey dress pants.
Barack Obama wearing a grey hoodie and black sweatpants.
Barack Obama wearing a yellow rain jacket and rain boots.
Barack Obama wearing a light blue dress shirt and navy blue suit.
Barack Obama wearing a black turtleneck and khaki pants.
Barack Obama wearing a red plaid shirt and jeans.
Barack Obama wearing a grey suit and a red tie.
Barack Obama wearing a white polo shirt and black shorts.
Barack Obama wearing a blue suit and a white dress shirt.
Barack Obama wearing a black t-shirt and black jeans.
Barack Obama wearing a purple dress shirt and black suit.
Barack Obama wearing a brown leather jacket and jeans.
Barack Obama wearing a white t-shirt and cargo pants.
Barack Obama wearing a grey suit and a blue tie.
Barack Obama wearing a black hoodie and black sweatpants.
Barack Obama wearing a green sweater and khaki pants.
Barack Obama wearing a navy blue suit and a white dress shirt.
Barack Obama wearing a red polo shirt and khaki shorts.
Barack Obama wearing a brown leather bomber jacket and jeans.
Barack Obama wearing a white dress shirt and black suit pants.
Barack Obama wearing a green camouflage jacket and cargo pants.
Barack Obama wearing a grey sweater and black dress pants.
Barack Obama wearing a blue plaid shirt and jeans.
Barack Obama wearing a black suit and a light blue dress shirt.

\end{verbatim} 
}

\end{appendices}

\end{document}